\documentclass[10pt,twocolumn,letterpaper]{article}

\usepackage[pagenumbers]{cvpr} % To force page numbers, e.g. for an arXiv version
\usepackage[accsupp]{axessibility}

\usepackage[pdftex]{graphicx}
\usepackage{amsmath}
\usepackage{amsfonts}
\usepackage{amsthm}
\usepackage{mathtools}
\usepackage{amssymb}
\usepackage{booktabs}
\usepackage{color}
\usepackage{bm}
\usepackage{multirow}
\usepackage{enumitem}
\usepackage{algorithm}
\usepackage{algorithmic}
\usepackage{mathcomp}

\newtheorem{remark}{Remark}
\newtheorem{definition}{Definition}
\newtheorem{theorem}{Theorem}

\usepackage{pifont}
\newcommand{\cmark}{\ding{51}}
\newcommand{\xmark}{\ding{55}}

\usepackage[pagebackref,breaklinks,colorlinks]{hyperref}

\usepackage[capitalize]{cleveref}
\crefname{section}{Sec.}{Secs.}
\Crefname{section}{Section}{Sections}
\Crefname{table}{Table}{Tables}
\crefname{table}{Tab.}{Tabs.}

\def\cvprPaperID{7443} % *** Enter the CVPR Paper ID here
\def\confName{CVPR}
\def\confYear{2023}

\begin{document}

%%%%%%%%% TITLE - PLEASE UPDATE
\title{Deep Curvilinear Editing: Commutative and Nonlinear Image Manipulation\\ for Pretrained Deep Generative Model}

\author{Takehiro Aoshima, Takashi Matsubara\\
Osaka University\\
1-3 Machikaneyama, Toyonaka, Osaka, 560-8531 Japan.\\
{\tt\small aoshima@hopf.sys.es.osaka-u.ac.jp, matsubara@sys.es.osaka-u.ac.jp}
% For a paper whose authors are all at the same institution,
% omit the following lines up until the closing ``}''.
% Additional authors and addresses can be added with ``\and'',
% just like the second author.
% To save space, use either the email address or home page, not both
}
\maketitle

%%%%%%%%% ABSTRACT
\begin{abstract}
  Semantic editing of images is the fundamental goal of computer vision.
  Although deep learning methods, such as generative adversarial networks (GANs), are capable of producing high-quality images, they often do not have an inherent way of editing generated images semantically.
  Recent studies have investigated a way of manipulating the latent variable to determine the images to be generated.
  However, methods that assume linear semantic arithmetic have certain limitations in terms of the quality of image editing, whereas methods that discover nonlinear semantic pathways provide non-commutative editing, which is inconsistent when applied in different orders.
  This study proposes a novel method called deep curvilinear editing (DeCurvEd) to determine semantic commuting vector fields on the latent space.
  We theoretically demonstrate that owing to commutativity, the editing of multiple attributes depends only on the quantities and not on the order.
  Furthermore, we experimentally demonstrate that compared to previous methods, the nonlinear and commutative nature of DeCurvEd facilitates the disentanglement of image attributes and provides higher-quality editing.
  % We have experimentally demonstrate the
\end{abstract}

%%%%%%%%% BODY TEXT
\section{Introduction}\label{sec:intro}
Generating and editing realistic images is one of the fundamental goals in computer vision.
Generative adversarial networks~(GANs)~\cite{Goodfellow2014} have emerged as a major image generation approach owing to the quality of generated images~\cite{Karras2018,Karras2019stylegan,Karras2019stylegan2,Miyato2018,Brock2019LargeSG,Karras2021} and provide various real-world applications~\cite{Choi2020StarGANVD,Wang2018VideotoVideoS,Zhu2017UnpairedIT,Schlegl2017UnsupervisedAD,FRIDADAR2018321,Abdal2021CLIP2StyleGANUE}.
Other notable methods include variational autoencoders (VAEs)~\cite{Higgins2017a}, conditional PixelCNN~\cite{Oord2016b}, and diffusion-based models~\cite{Ho2020,Song2020}, which are collectively called deep generative models.
However, deep generative models cannot inherently edit images semantically.
They can be viewed as mappings from latent space to image space, and the latent variables determine the generated images.
Therefore, several methods have been developed to train deep generative models such that each semantic attribute the user wants to edit is assigned to each element of the latent variable~\cite{infogan,Liu2020OOGANDG,Lin2020InfoGANCRAM} (see the first column of Table~\ref{tab:compare-table}).
However, this approach requires computationally expensive training and can conflict with the quality of image generation.
Other studies have developed image-to-image translation that translates images from one domain to another~\cite{Isola2017ImagetoImageTW,Zhu2017UnpairedIT,Wu2019RelGANMI}.
However, this approach also requires training from scratch and limits image editing to be discontinuous unless combined with latent variable manipulation.
Therefore, a general and promising approach is necessary to identify manipulations on latent variables of already trained models that edit images semantically.
% ~\cite{gansteerability,Antioine2020,GANalyze,Shen2020InterpretingTL,ZhuangICLR2021,Haas2022TensorbasedEE,voynov2020unsupervised,2020ganspace,shen2021closedform,Tzelepis_2021_ICCV,Choi2021DoNE,Ramesh2018ASR,Abdal2021,Liang2021SSFlowSN,tewari2020stylerig,chen2022,Khrulkov2021LatentTV}.

\begin{table*}[t]
  \centering
  \footnotesize
  \caption{Comparison of Our Proposal against Related Methods.}
  \label{tab:compare-table}
  \begin{tabular}{lcccc}
    \toprule
                                         & Training under constraints                            & Linear arithmetic                                             & Vector fields/Local basis                               & DeCurvEd                                                \\
                                         & \cite{infogan,Liu2020OOGANDG,Lin2020InfoGANCRAM}      & \cite{2020ganspace,shen2021closedform,voynov2020unsupervised} & \cite{Choi2021DoNE,Ramesh2018ASR,Tzelepis_2021_ICCV}    & (proposed)                                              \\
    \midrule
    Global coordinate                    & Cartesian                                             & oblique                                                       & (only local)                                            & curvilinear                                             \\
    No retraining                        & \textcolor{red}{\xmark}                               & \textcolor{green}{\cmark}                                     & \textcolor{green}{\cmark}                               & \textcolor{green}{\cmark}                               \\
    Nonlinear edit                       & --                                                    & \textcolor{red}{\xmark}                                       & \textcolor{green}{\cmark}                               & \textcolor{green}{\cmark}                               \\
    Commutative edit                     & \textcolor{green}{\cmark}                             & \textcolor{green}{\cmark}                                     & \textcolor{red}{\xmark}                                 & \textcolor{green}{\cmark}                               \\
    \midrule
    \raisebox{1.4cm}{Conceptual diagram} & \includegraphics*[scale=0.65]{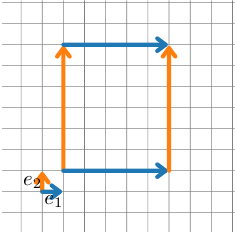} & \includegraphics*[scale=0.65]{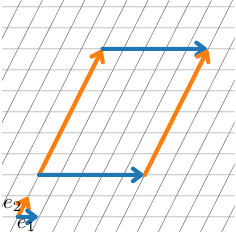}            & \includegraphics*[scale=0.65]{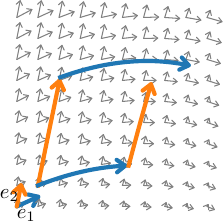} & \includegraphics*[scale=0.65]{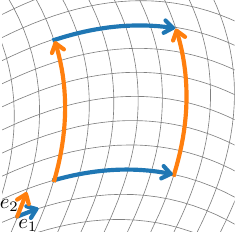} \\
    \bottomrule
  \end{tabular}
  \vspace*{-2mm}
\end{table*}

A study reported that adding certain vectors to latent variables can modify the corresponding attributes of the object in the generated images~\cite{Radford2016UnsupervisedRL}.
This indicates that the latent space can be regarded as a linear vector space.
Some studies have aimed to identify attribute vectors in a supervised or an unsupervised manner~\cite{GANalyze,Shen2020InterpretingTL,Antioine2020,Eliezer2021,ZhuangICLR2021,Haas2022TensorbasedEE,voynov2020unsupervised,2020ganspace,shen2021closedform,Oldfield2021TensorCA}.
In any case, these studies introduce the strong assumption of linear semantic arithmetic on the latent space (see the second column of Table~\ref{tab:compare-table}), which limits the quality of image editing.
Other studies have proposed methods to determine attribute vectors depending on the position in the latent space, that is, the attribute vector fields (or local attribute coordinates).~\cite{Tzelepis_2021_ICCV,Choi2021DoNE,Ramesh2018ASR,Abdal2021,Liang2021SSFlowSN,tewari2020stylerig,chen2022,Khrulkov2021LatentTV,gansteerability,Eliezer2021}; these methods edit an image attribute by moving the latent variable nonlinearly along the corresponding attribute vector field.
% Supervised linear: \cite{GANalyze,Shen2020InterpretingTL,ZhuangICLR2021}
% Unsupervised linear: \cite{voynov2020unsupervised,2020ganspace,shen2021closedform}
% Supervised multilinear: \cite{Haas2022TensorbasedEE}
% Unsupervised multilinear: \cite{Oldfield2021TensorCA}
% Self-supervised linear: \cite{Antioine2020}
% Self-supervised nonlinear: \cite{gansteerability,Eliezer2021}
% Supervised nonlinear methods: \cite{chen2022,Abdal2021,Liang2021SSFlowSN,tewari2020stylerig,Khrulkov2021LatentTV}
% Unsupervised nolinear methods: \cite{Tzelepis_2021_ICCV,Choi2021DoNE,Ramesh2018ASR}
Although this approach is elegant, the edits of different attributes are generally non-commutative (see the third column of Table~\ref{tab:compare-table}).
That is, what we get is different when we edit attributes (denoted by $e_1$ and $e_2$) one after another, or in the reverse order.
This property can harm the disentanglement between attributes considering that the relationships among attributes change at different points.
In contrast, linear arithmetic ensures that the edits of different attributes are commutative.

To overcome this dilemma, this study proposes \emph{deep curvilinear editing (DeCurvEd)}, a method that determines a set of commuting attribute vector fields in the latent space of a pre-trained deep generative model.
%  in an unsupervised manner.
The key idea is adopted from the theorem that a set of vector fields is locally expressed as partial derivatives of a coordinate chart if it is linearly independent and commuting~\cite{Lee2012}.
Therefore, we define a curvilinear coordinate system globally~\cite{Arfken2012} by employing a normalizing flow~\cite{Kingma2018GlowGF,chen2018neuralode}, from which we derive commuting vector fields (see the rightmost panel of Table~\ref{tab:compare-table}).
% For evaluation, we propose \emph{CurvilinearGANSpace} by combining DeCurvEd with GANs and experimentally demonstrate that its editing is commutative and high-quality.
The advantages of DeCurvEd are as follows (see also Table~\ref{tab:compare-table}):
\begin{enumerate}[itemsep=0ex]
  \item Edits of different attributes are always commutative, unlike previous methods that assume attribute vector fields (e.g.,~\cite{Tzelepis_2021_ICCV}).
        Therefore, an edited image does not depend on the order of editing, but on the amount of editing performed.
  \item Edits are nonlinear, which indicates that DeCurvEd provides better editing quality than methods that assume linear arithmetic of attribute vectors (e.g.,~\cite{voynov2020unsupervised}).
  \item DeCurvEd does not require retraining of deep generative models, but identifies attribute vector fields in the latent space of pre-trained models, unlike image-to-image translation and training under constraints (e.g.,~\cite{infogan,Isola2017ImagetoImageTW}).
  \item We propose \emph{CurvilinearGANSpace} by combining DeCurvEd with GANs and experimentally demonstrate that the nonlinear and commutative nature disentangles attributes and enables high-quality editing.
  \item The key idea is not limited to GANs, and is available for any generative models that are conditioned on latent variables, including VAEs~\cite{Higgins2017a}, conditional PixelCNN~\cite{Oord2016b}, and diffusion-based models~\cite{Ho2020,Song2020}.
\end{enumerate}

\section{Related Work}\label{sec:related}
\paragraph{Image Editing by Deep Generative Models}
Most generative models define a mapping that maps a latent variable to a data sample (that is, an image in this study).
Previous studies on deep learning-based generative models have confirmed that manipulating the latent variable can determine the image to be generated~\cite{Higgins2017a,Goodfellow2014}.
% Following studies have attempted to identify the manipulation that edits the image as desired~\cite{Higgins2017betaVAELB,Bose2021NeurIntL,Radford2016UnsupervisedRL}.
Radford et al.~\cite{Radford2016UnsupervisedRL} identified a semantically meaningful vector in the latent space based on the difference between two groups.
% In particular, they defined an attribute vector (e.g., a vector of ``wearing sunglasses'') as the difference between the average latent variable of the corresponding group (e.g., images of persons with sunglasses) and that of the remaining (e.g., images of persons without sunglasses).
% Then, one can make an image have the attribute (e.g., make a person wear sunglasses) by adding the attribute vector to the image's latent variable.
Then, an attribute can be imposed on an image by adding the attribute vector to the latent variable of the image.
This discovery attracted wide attention and spurred research into semantic image editing by manipulating the latent variables.

Several studies have developed methods for training models under constraints to easily determine attribute vectors, rather than identifying attribute vectors after training~\cite{Higgins2017betaVAELB,infogan,Lin2020InfoGANCRAM,Liu2020OOGANDG}.
These methods often make each element of the latent variable as independent as possible~\cite{Higgins2017betaVAELB}.
Then, one element is assigned to a group (an attribute) that changes collectively and is independent of other groups in the image~\cite{Bengio2013RepresentationLA}.
These methods can be viewed as introducing a Cartesian coordinate system to the latent space and assigning one attribute to each axis.
However, constraints often conflict with other training criteria (such as likelihood and Gaussian prior) and result in models with inferior quality and diversity.
Additionally, the generated models needed to be trained from scratch, which incurs computation costs.
See the column ``Training under constraints'' in Table~\ref{tab:compare-table}.

Moreover, some studies have developed image-to-image transformation, which maps an image from one domain to another rather than manipulating a latent variable~\cite{Isola2017ImagetoImageTW,Zhu2017UnpairedIT,Wu2019RelGANMI}.
This approach limits image editing only between domains or needs to be combined with a latent space.

\paragraph{Discovering Linear Attribute Arithmetic}
Several studies have investigated linear manipulations on latent variables in the already trained deep generative models~\cite{voynov2020unsupervised,2020ganspace}.

SeFa and related methods attempted to find semantic directions by analyzing the weight parameters~\cite{shen2021closedform,Zhu2021LowRankSI,zhu2022resefa}.
Voynov and Babenko~\cite{voynov2020unsupervised} proposed an unsupervised framework that learns a set of semantic directions.
According to this framework, changing a latent variable along a semantic direction will edit one attribute of the corresponding image, and the degree of change in the attribute will be proportional to the amount of change in the latent variable.
GANSpace~\cite{2020ganspace} applied a principal component analysis (PCA) to extract a lower-dimensional subspace of the latent space, assuming each principal component corresponds to an attribute.
These methods assume linear arithmetic of the attribute vectors; that is, they introduce an oblique coordinate system to the latent space.
See the column ``Linear arithmetic'' in Table~\ref{tab:compare-table}.

Because the distribution of real-world data is often biased and skewed, it is unlikely that the latent space is flat and homogeneous.
Khrulkov et al.~\cite{Khrulkov2021LatentTV} found that different directions correspond to the same attribute at different locations in the latent space.
Therefore, the above methods are limited in terms of image editing quality.

\paragraph{Discovering Semantic Vector Fields}
The direction corresponding to an attribute varies depending on the location in the latent space, thereby indicating that a set of directions corresponding to attributes forms a set of vector fields, rather than linear arithmetic.
If so, one can edit an attribute of an image by moving the latent variable nonlinearly along the vector field corresponding to the attribute instead of adding an attribute vector.
Tzelepis et al.~\cite{Tzelepis_2021_ICCV} proposed WarpedGANSpace, which learns a set of vector fields, each of which is defined as a gradient flow of an RBF function on the latent space.
Choi et al.~\cite{Choi2021DoNE} learned a local basis at every point of the latent space such that each element of the local basis corresponds to an attribute.
StyleFlow~\cite{Abdal2021} and SSFlow~\cite{Liang2021SSFlowSN} used normalizing flows to define a local coordinate system.
On an $N$-dimensional manifold, a local basis, local coordinate system, and a set of $N$ linearly independent vector fields are compatible; such vector fields are called coordinate vector fields (see Example 8.2, \cite{Lee2012}).
However, because the coordinate system in the above studies is defined only locally, multiple edits may be inconsistent globally.
We will demonstrate this in the following section.
See the column ``Vector fields/Local basis'' in Table~\ref{tab:compare-table}.

Some studies have attempted to define a (Riemannian) metric on the latent space~\cite{Arvanitidis2018,Arvanitidis2020,Chen2018g}.
These methods successfully interpolate between two images by nonlinearly moving latent variables along the geodesic; however, they are insufficient for attribute editing.
Some others attempted complex and dynamic editing specified by text rather than attributes~\cite{Tzelepis2022}; nevertheless, such methods cannot be directly compared to ours.

\begin{figure*}[t]
  \centering
  \includegraphics[scale=0.35]{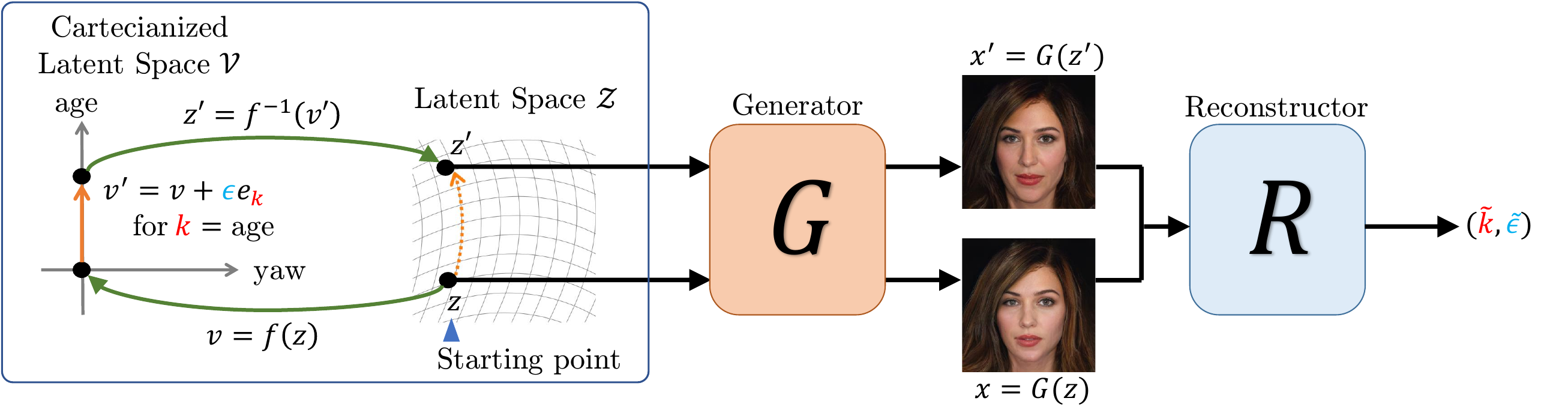}
  \vspace*{-3mm}
  \caption{The concept diagram.
    The left part shows how DeCurvEd edits an attribute.
    The right part shows its combination with GANs called CurvilinearGANSpace.
  }
  \label{fig:framework}
  \vspace*{-2mm}
\end{figure*}

\section{Theoretical Background}
We introduce the theoretical background of the proposed and related methods introduced in Section~\ref{sec:related}.
Theorems in this paper are basic knowledge about manifolds; readers unfamiliar with this topic are referred to the reference~\cite{Lee2012}.
Remarks are our findings.

Let $\mathcal X$ and $\mathcal Z$ denote an image space and a latent space of a deep generative model, respectively.
The generator $G$ (also called decoder) of the deep generative model is a mapping from the latent space $\mathcal Z$ to the image space $\mathcal X$; given a latent variable $z\in\mathcal Z$, the generator produces an image $x\in\mathcal X$ as $x=G(z)$.
We assume the latent space $\mathcal Z$ to be an $N$-dimensional space diffeomorphic to a Euclidean space.

Let $\{z^i\}_{i=1}^{N}$ denote the coordinate system (i.e., the basis) on a neighborhood of the point $z\in\mathcal Z$.
Let $Z_k$ denote a vector field on the latent space $\mathcal Z$ indexed by $k$, that is, $Z_k:\mathcal Z\rightarrow\mathcal T_z\mathcal Z$, where $\mathcal T_z\mathcal Z$ is the tangent space (i.e., the space of tangent vectors or velocities) of the latent space $\mathcal Z$ at point $z$.
At point $z$, the coordinate system on tangent space $\mathcal T_z\mathcal Z$ is denoted by $\{\frac{\partial}{\partial z^i}\}_{i=1}^{N}$, and a vector field $Z_k$ is expressed as $Z_k(z)=\sum_{i=1}^N Z_k^i(z)\frac{\partial}{\partial z^i}$ for smooth functions $Z_k^i:\mathcal Z\rightarrow\mathbb R$.
A method that assumes attribute vector fields~\cite{Tzelepis_2021_ICCV,Choi2021DoNE,Ramesh2018ASR} edits an attribute $k$ of an image $x$ by integrating a latent variable $z$ along the vector field $Z_k$ that corresponds to attribute $k$; the edited image is $x'=G(z')$ for
\begin{equation}
  \textstyle z'=z+\int_0^t Z_k(z(\tau))\mathrm{d}\tau, \label{eq:integral_curve}
\end{equation}
where $z(0)=z$, and $t\in\mathbb R$ denotes the change amount of attribute $k$.
$t$ may be positive or negative.
We rewrite the above equation using a \emph{flow}, denoted by $\phi_k^t:\mathcal Z\rightarrow\mathcal Z$ for $t$ as:
\begin{equation}
  z'=\phi^t_k(z).
\end{equation}
Then, we define the commutativity of editing as follows:
\begin{definition}[Commutativity]
  Edits of attributes $k$ and $l$ are commutative if and only if the corresponding flows $\phi_k^t$ and $\phi_l^s$ are commuting, that is, it holds that $\phi_l^s\circ\phi_k^t=\phi_k^t\circ\phi_l^s$ for any $s,t\in\mathbb R$ at any point $z$ on the latent space $\mathcal Z$.
\end{definition}
Intuitively, making a person smile and then wear sunglasses results in the same image as making the person wear sunglasses and then smile if the vector fields corresponding to smiling and wearing sunglasses are commuting.
Else, edits in different orders produce different images.

\begin{remark}\label{remark:linear_is_commutative}
  A method that assumes linear attribute arithmetic (e.g.,~\cite{2020ganspace,shen2021closedform,voynov2020unsupervised}) is a special case of a method that assumes attribute vector fields, and its edits are commutative.
\end{remark}

% This is because attribute vector fields independent of position provide linear attribute arithmetic.
See Appendix~\ref{appendix:proof_of_remarks} for formal proofs of any remarks in this manuscript.
Therefore, we can discuss a method that assumes linear attribute arithmetic in the same context.
We introduce the following theorem.

\begin{theorem}[Commuting Vector Fields, Theorem 9.44, \cite{Lee2012}]\label{theorem:commuting_vector_fields}
  Two flows $\phi_k$ and $\phi_l$ are commuting if and only if the underlying vector fields $Z_k$ and $Z_l$ are commuting.
\end{theorem}

This theorem suggests the following remark:

\begin{remark}\label{remark:vectorfields_are_not_commuting}
  In general, edits by a method that assumes attribute vector fields (e.g.,~\cite{Choi2021DoNE,Ramesh2018ASR,Tzelepis_2021_ICCV}) are non-commutative.
\end{remark}

In addition, we introduce the following theorem.

\begin{theorem}[Canonical Forms for Commuting Vector Fields, Theorem 9.46, \cite{Lee2012}]\label{theorem:commutativeVF}
  Let vector fields $Z_1,Z_2,\dots,Z_N$ on an $N$-dimensional space $\mathcal Z$ be linearly independent and commuting on an open set $\mathcal U\subset\mathcal Z$.
  At each $z\in\mathcal U$, there exists a smooth coordinate chart $\{\frac{\partial}{\partial s^i}\}_{i=1}^N$ centered at $z$ such that $\frac{\partial}{\partial s^i}=Z_i$.
\end{theorem}
Furthermore, given a smooth coordinate chart $\{\frac{\partial}{\partial s^i}\}_{i=1}^N$, vector fields $Z_i=\frac{\partial}{\partial s^i}$ are commuting.
A coordinate chart is a nonlinear bijective mapping to Euclidean space.
Therefore, intuitively, a set of $N$ linearly independent and commuting vector fields on an $N$-dimensional space $\mathcal Z$ is equivalent to a set of $N$ vector fields along the axes of a coordinate system up to geometric transformation.

\section{Method}
\subsection{DeCurvEd}
Given the theoretical background, we propose DeCurvEd, as shown in Fig.~\ref{fig:framework}.
% We define a set of commutative attribute vector fields on the latent space $\mathcal Z$ by coordinate transformation.
Intuitively, we consider the case where the open set $\mathcal U$ in Theorem~\ref{theorem:commutativeVF} is not a proper subset but equal to the latent space $\mathcal Z$.

We prepare an $N$-dimensional Euclidean space $\mathcal V$ and call it the Cartesianized latent space, whose coordinate system $\{v^i\}_{i=1}^N$ is a global Cartesian coordinate system.
Let $e_k$ denote the $k$-th element of the standard basis of the tangent space, that is, $e_k\coloneqq \frac{\partial}{\partial v^k}$.
Then, the vector filed $\tilde Z_k$ corresponding to attribute $k$ is defined as
\begin{equation}
  \tilde Z_k\coloneqq e_k.
\end{equation}
As discussed in the previous section, vector fields $\tilde Z_k$ and $\tilde Z_l$ defined in this way are commuting for any $k$ and $l$.
The flow $\psi^t_k:\mathcal V\rightarrow\mathcal V$ that arises from the vector filed $\tilde Z_k$ is given by
\begin{equation}
  \textstyle\psi^t_k(v)\coloneqq v+\int_0^t e_k\mathrm{d}\tau=v+t\,e_k.
\end{equation}
The flows $\psi_k$ are commuting because $(\psi^s_l\circ \psi^t_k)(v)=v+t\,e_k+s\,e_l=(\psi^t_k\circ \psi^s_l)(v)$.
We introduce a smooth bijective mapping $f:\mathcal Z\rightarrow\mathcal V,z\mapsto v$ that corresponds to the coordinate chart in Theorem~\ref{theorem:commutativeVF}.
The mapping $f$ can be implemented using a normalizing flow; however, it is not limited to~\cite{Kingma2018GlowGF,chen2018neuralode,grathwohl2019ffjord}.
We define a flow $\phi_k$ that edits attribute $k$ on the latent space $\mathcal Z$ as:
\begin{equation}
  \phi_k^t\coloneqq f^{-1}\circ\psi_k^t\circ f.
\end{equation}
See also the left half of Fig.~\ref{fig:framework}.
We redefine the edit as Algorithm~\ref{alg:edit} in Appendix~\ref{appendix:algorithm}.

Subsequently, one can generate an edited image $x'=G(z')$ using generator $G$.
Deep generative models such as GANs do not have an inherent way of inferring a latent variable $z$ from an image $x$; this is outside the scope of this study.
Interested readers are can refer to this survey~\cite{Xia2022}.

\subsection{Theoretical Analysis}
The pushforward $f_*$ is a mapping naturally induced by the mapping $f$, which maps a tangent vector (or a basis) on the latent space $\mathcal Z$ to that on the Cartesianized latent space $\mathcal V$.
Also, the pushforward $(f^{-1})_*$ maps the Cartesian coordinate system on the Cartesianized latent space $\mathcal V$ and implicitly defines a coordinate system on the latent space $\mathcal Z$~\cite{Lee2012}.
A coordinate system defined by a bijective transformation of a Cartesian coordinate is called a curvilinear coordinate~\cite{Arfken2012}.
Therefore, we name this method deep curvilinear editing (DeCurvEd).
Because the mapping $f$ is defined globally between spaces $\mathcal Z$ and $\mathcal V$, the curvilinear coordinate system is also defined globally.
The pushforward $(f^{-1})_*$ can define commuting vector fields on $\mathcal Z$ by push-forwarding the coordinate vector fields on $\mathcal V$.
Therefore, we make the following remarks.

\begin{remark}\label{remark:nonlinear_commutative}
  Using DeCurvEd, any edits of attributes in the latent space $\mathcal Z$ can be nonlinear and commutative.
\end{remark}

\begin{remark}\label{remark:proposed_is_VF}
  DeCurvEd can define vector fields on the latent space $\mathcal Z$ and is a special case of a method that assumes attribute vector fields (e.g.,~\cite{Choi2021DoNE,Ramesh2018ASR,Tzelepis_2021_ICCV}).
\end{remark}

\begin{remark}\label{remark:linear_is_proposed}
  A method that assumes linear attribute arithmetic (e.g.,~\cite{2020ganspace,shen2021closedform,voynov2020unsupervised}) is a special case of DeCurvEd, with a linear mapping $f$.
\end{remark}

Therefore, DeCurvEd enjoys the advantages of both attribute arithmetic and vector fields.
All theories and remarks are not dependent on the properties of particular models.
Thus, we make the following remark.

\begin{remark}\label{remark:generality}
  DeCurvEd offers attribute editing for any generative models conditioned on latent variables, including GANs~\cite{Goodfellow2014}, VAEs (see Fig.~4 of \cite{Higgins2017a}), conditional PixelCNN~\cite{Oord2016b}, and diffusion-based models (see Fig.~8 of \cite{Ho2020} and Fig.~4 of \cite{Song2020}).
\end{remark}

\subsection{CurvilinearGANSpace}\label{sec:VoynovAndBabenko}
Attribute editing by DeCurvEd is available for any deep generative models and for both supervised and unsupervised learning.
This study adopted the unsupervised training framework for GANs proposed by Voynov and Babenko~\cite{voynov2020unsupervised}, as shown in the right half of Fig.~\ref{fig:framework}.
Following previous studies, we call it \emph{CurvilinearGANSpace}.

Given a latent variable $z$, CurvilinearGANSpace randomly edits index $k$ by $\epsilon$ and produces an edited one $z'$.
In some cases, only the first $N'$ indices of all $N$ indices are candidates for editing.
We prepare a neural network called reconstructor $R$, which accepts the pair of generated images $x=G(z)$ and $x'=G(z)$ and regresses the edited index $k$ and the change amount $\epsilon$.
In particular, one output is an $N'$-dimensional vector $\hat k$ to regress the edited index $k$: the loss function is the classification error $L_\mathrm{cls}(k,\hat k)$, which is defined as the cross-entropy.
As the mapping $f$ minimizes this error, image editing of index $k$ becomes easier for the reconstructor $R$ to distinguish from image editing of other indices $l\neq k$, thereby assigning one attribute to each vector field and facilitating the disentanglement between attributes.
The other output $\hat\epsilon$ is a scalar regressing the change amount $\epsilon$; the loss function is the regression error $L_\mathrm{reg}(\epsilon,\hat\epsilon)$ defined as the absolute error.
As this error is minimized, the change in the latent variable continuously matches the semantic change in the image.

Additionally, we introduce a regularization term $L_\mathrm{nl}$ to be minimized for the mapping $f$;
\begin{equation}
  \textstyle L_\mathrm{nl}(z)=(\log{\det{|\frac{\partial f}{\partial z}|}})^2.
\end{equation}
The Jacobian determinant $\det\frac{\partial f}{\partial z}$ of the mapping $f$ indicates the extent to which the latent space $\mathcal Z$ is stretched by the mapping $f$; when it is 1.0, the mapping $f$ is isometric.
Subsequently, this term $L_\mathrm{nl}$ avoids extreme deformation of the latent space $\mathcal Z$ by the mapping $f$.
The final objective function is defined as:
\begin{equation}
  \textstyle \mathbb{E}_{z, k, \epsilon} \left[L_\mathrm{cls}(k, \hat{k}) + \lambda L_\mathrm{reg} (\epsilon, \hat{\epsilon}) + \alpha L_\mathrm{nl}(z)\right],\label{eq:loss}
\end{equation}
where $\lambda,\alpha\in\mathbb R$ are hyperparameters weighing objectives.
See also Algorithm~\ref{alg:training} in Appendix~\ref{appendix:algorithm} for more details.

\begin{table}[t]
  \caption{Datasets and Settings.}\label{tab:datasets}
  \centering
  \footnotesize
  \tabcolsep=2mm
  \begin{tabular}{lllcc}
    \toprule
    \textbf{Dataset}                       & \textbf{GANs}                        & \textbf{Reconstructor}        & $N$ & $N'$ \\
    \midrule
    MNIST~\cite{LeCun2005TheMD}            & SNGAN~\cite{Miyato2018}              & LeNet~\cite{726791}           & 128 & 128  \\
    AnimeFaces~\cite{Jin2017TowardsTH}     & SNGAN~\cite{Miyato2018}              & LeNet~\cite{726791}           & 128 & 128  \\
    ILSVRC~\cite{Deng2009ImageNetAL}       & BigGAN~\cite{Brock2019LargeSG}       & ResNet-18~\cite{He2016DeepRL} & 120 & 120  \\
    CelebA-HQ~\cite{Liu2015DeepLF}         & ProgGAN~\cite{Karras2018}            & ResNet-18~\cite{He2016DeepRL} & 512 & 200  \\
    CelebA-HQ~\cite{Liu2015DeepLF}         & StyleGAN2~\cite{Karras2019stylegan2} & ResNet-18~\cite{He2016DeepRL} & 512 & 200  \\
    LSUN Car~\cite{KrambergerPotocnik2020} & StyleGAN2~\cite{Karras2019stylegan2} & ResNet-18~\cite{He2016DeepRL} & 512 & 200  \\
    \bottomrule
  \end{tabular}
  \vspace*{-2mm}
\end{table}

\section{Experiments}
\subsection{Experimental Setting}
\paragraph{Datasets, Backbones, and Comparison Methods}
We examined CurvilinearGANSpace and related methods using combinations of datasets, GANs, and reconstructors, as summarized in Table~\ref{tab:datasets}.
$N$ denotes the number of dimensions of the latent space $\mathcal Z$, and $N'$ denotes the number of dimensions used for training.
For StyleGAN2, $W$ space was used as the latent space.
For ILSVRC and CelebA-HQ, we used pre-trained models from their official repositories.
These experimental settings are identical to those in previous studies~\cite{voynov2020unsupervised,Tzelepis_2021_ICCV}.
See Appendix~\ref{appendix:datasets} and the references~\cite{voynov2020unsupervised,Tzelepis_2021_ICCV} for more details.

For comparison, we used a method that assumes linear arithmetic~\cite{voynov2020unsupervised} and a method that assumes attribute vector fields called WarpedGANSpace~\cite{Tzelepis_2021_ICCV}.
To clarify the difference, we hereafter refer to the former method as LinearGANSpace.
We used their pre-trained models for all but the LSUN Car dataset and used our own trained models for the LSUN Car dataset, each trained in the same framework.
\footnote{\label{footnote:linear}\url{https://github.com/anvoynov/GANLatentDiscovery} for LinearGANSpace.}
\footnote{\label{footnote:warpd}\url{https://github.com/chi0tzp/WarpedGANSpace} for WarpedGANSpace and attribute predictors.}

\paragraph{Architectures and Hyperparameters}
As the bijective mapping $f$, we used a continuous normalizing flow with six concatsquash layers~\cite{grathwohl2019ffjord}.
We set the number of hidden units equal to the input dimension and used hyperbolic tangent function as its activation function.
See Appendix~\ref{appendix:implementation} for more introduction.
We used Adam optimizer~\cite{Kingma2015} with a constant learning rate of $10^{-4}$.
We used $\lambda=0.25$, which is equivalent to that used by previous studies~\cite{voynov2020unsupervised,Tzelepis_2021_ICCV}.
For simplicity, we used $\alpha=1$.

\begin{figure*}[t]
  \centering
  \tabcolsep=.5mm
  \begin{minipage}{100mm}
    \begin{tabular}{cl|c}
      \rotatebox{90}{\footnotesize \ \ Linear} & \includegraphics[clip, width=8.1cm]{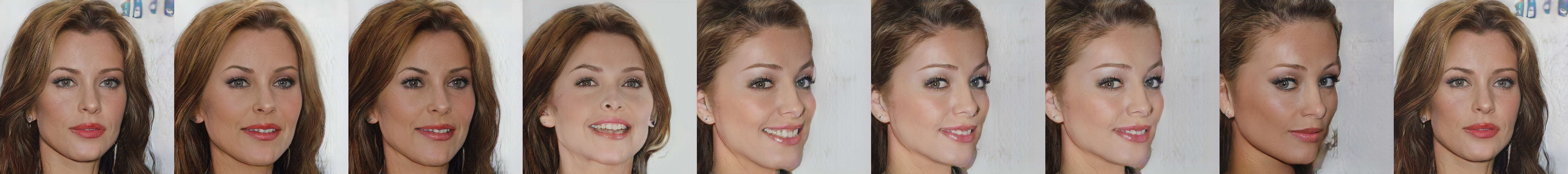}                                                                                              & \includegraphics[clip, width=0.9cm]{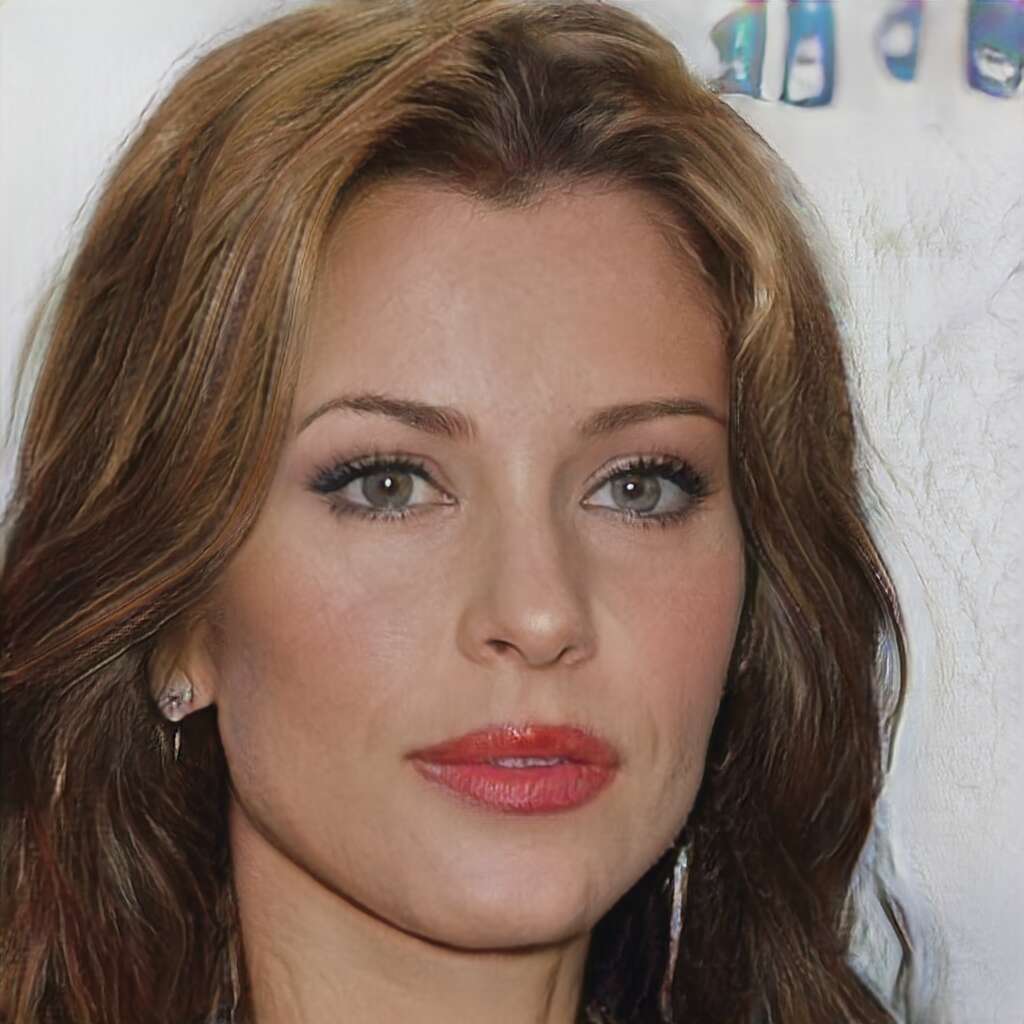} \\[-.8mm]
      \rotatebox{90}{\footnotesize Warped}     & \includegraphics[clip, width=8.1cm]{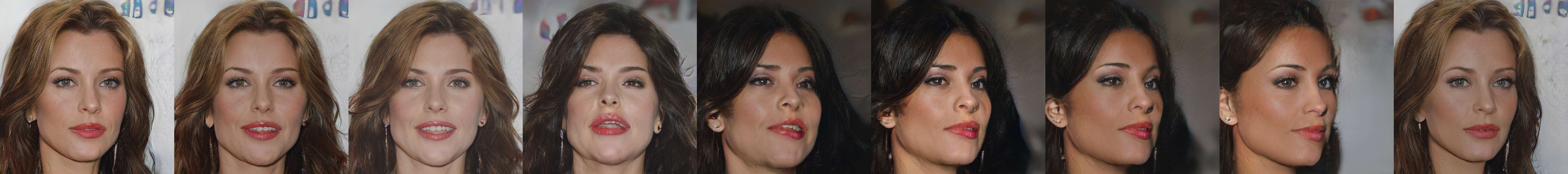}                                                                                              & \includegraphics[clip, width=0.9cm]{./fig/commutative_woman1_proggan_original.jpg} \\[-.8mm]
      \rotatebox{90}{\footnotesize \ \ \ Ours} & \includegraphics[clip, width=8.1cm]{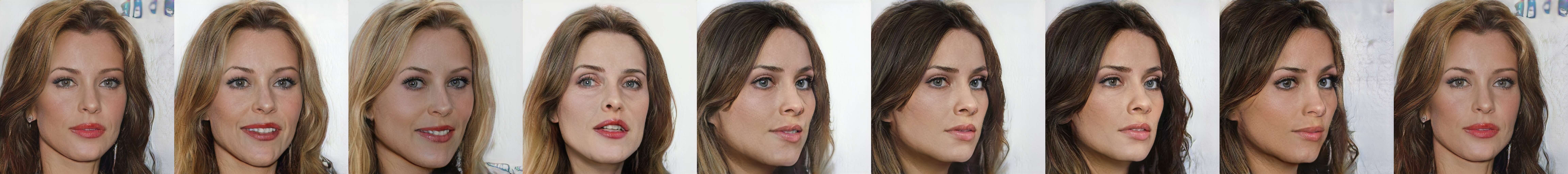}                                                                                            & \includegraphics[clip, width=0.9cm]{./fig/commutative_woman1_proggan_original.jpg} \\[-1mm]
                                               & \footnotesize \hspace{3.3mm}O\hspace{5.5mm}$+$S\hspace{5.0mm}$+$B\hspace{5.0mm}$+$P\hspace{5.0mm}$+$Y\hspace{5.0mm}$-$S\hspace{5.0mm}$-$B\hspace{5.0mm}$-$P\hspace{5.0mm}$-$Y & \footnotesize O
    \end{tabular}
  \end{minipage}\hspace*{3mm}
  \begin{minipage}{60mm}
    \begin{tabular}{cl|c}
      \rotatebox{90}{\footnotesize \ \ Linear} & \includegraphics[clip, width=4.5cm]{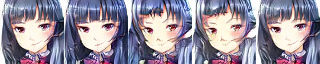}                              & \includegraphics[clip, width=0.9cm]{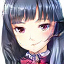} \\[-.8mm]
      \rotatebox{90}{\footnotesize Warped}     & \includegraphics[clip, width=4.5cm]{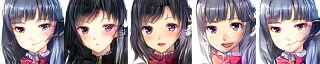}                              & \includegraphics[clip, width=0.9cm]{./fig/commutative_anime1_original.jpg} \\[-.8mm]
      \rotatebox{90}{\footnotesize \ \ \ Ours} & \includegraphics[clip, width=4.5cm]{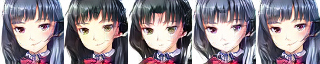}                            & \includegraphics[clip, width=0.9cm]{./fig/commutative_anime1_original.jpg} \\[-1mm]
                                               & \footnotesize \hspace{3.3mm}O\hspace{5.5mm}$+$C\hspace{5.0mm}$-$L\hspace{5.0mm}$-$C\hspace{5.0mm}$+$L & \footnotesize O
    \end{tabular}
  \end{minipage}
  \vspace*{-3mm}
  \caption{Results of sequential editing of attributes from left to right.
    (left) CelebA-HQ+ProgGAN.
    The change amount was set to $\tilde t=0.3$ for smile, bangs, and pitch, and $\tilde t=1.0$ for yaw.
    (right) AnimeFaces+SNGAN.
    Each row shows the results of LinearGANSpace, WarpedGANSpace, and CurvilinearGANSpace, from top to bottom.
    The signs $+$ and $-$ denote the addition and the subtraction of the corresponding attributes, respectively.
    O: original, S: ``smile'', B: ``bangs'', P: ``pitch'', Y: ``yaw''.
    C: ``hair color'', L: ``hair length''.
  }
  \label{fig:result-commutive}
  \vspace*{-2mm}
\end{figure*}

\paragraph{Evaluation Metrics}
For CelebA-HQ, we measured the attribute scores of generated images using separate pre-trained attribute predictors $A_k:\mathcal X\rightarrow\mathbb R$.\textsuperscript{\ref{footnote:warpd}}
FairFace measured age, gender, and race (skin color) attributes~\cite{karkkainenfairface}, and CelebA-HQ attributes classifier measured smile, beard, and bangs attributes~\cite{jiang2021talk} from 0 to 1; Hopenet measured face directions, yaw and pitch, in degree~\cite{Doosti_2020_CVPR}.
Additionally, ArcFace measured the identity score $I(\cdot,\cdot):\mathcal X\times\mathcal X\rightarrow[0,1]$ to evaluate whether two images are of the same person~\cite{deng2018arcface}.
We selected index $k$ as the one corresponding to that attribute if the measured attribute score has the largest covariance with the change amounts of index $k$.
The above procedure is identical to that of the previous study~\cite{Tzelepis_2021_ICCV}.

The amount $t$ by which attribute $k$ of latent variable $z$ is edited differs from the amount by which the corresponding attribute score $A_k$ of the generated image $x=G(x)$ is changed.
For a fair comparison, we normalized the change amount $t$ such that the measured attribute score changes by 5 degrees for the pitch and yaw attributes and 0.1 for others, and denoted the amount by $\tilde t=0.1$.

After index identification and normalization, we used several evaluation metrics.
We defined \emph{commutativity error} of attributes $k$+$l$ to evaluate how commutative the image editing is by measuring the difference in the attribute score between images with edits of attributes $k$ and $l$ applied in different orders.
We defined \emph{side effect error} to evaluate the disentanglement between attributes by measuring how much an edit of the target attribute $k$ changes the other attributes $l\neq k$ as undesired side effects.
We also defined \emph{identity error} to evaluate the disentanglement by measuring how much an edit of the target attribute $k$ reduces the identity score.
The errors in edits of multiple attributes are defined similarly.
% We used \emph{consistency error} of attributes $k$+$l$+$m$, proposed in \cite{Abdal2021}, to evaluate the editing quality.
% It is the error in the attribute score after an edit of attribute $l$ combined with an edit of one of the two remaining attributes $k$ and $m$.
% We divided the errors by the change amount $t$ of target attribute and showed them in percentages.

Owing to the availability of attribute predictors, these evaluations were performed only for CelebA-HQ.
See Appendix~\ref{appendix:metrics} for the detailed procedures and definition.
For other datasets, we manually selected the index $k$, following previous studies~\cite{voynov2020unsupervised,Tzelepis_2021_ICCV}.

\begin{table}[t]
  \centering\footnotesize
  \tabcolsep=1.5mm
  \caption{Commutativity Errors [\%] of StyleGAN2.}
  \label{tab:commutativity-stylegan2}
  \begin{tabular}{lcccccccccc}
    \toprule
                                                 & \multicolumn{1}{c}{\textbf{A+G}}                & \multicolumn{1}{c}{\textbf{R+P}}    & \multicolumn{1}{c}{\textbf{B+Y}}    \\
    \midrule
    LinearGANSpace~\cite{voynov2020unsupervised} & \hphantom{0}\textbf{0.01} / \textbf{0.05}       & \textbf{0.02}/ \textbf{0.07}        & \textbf{0.02} / \textbf{0.15}       \\
    WarpedGANSpace~\cite{Tzelepis_2021_ICCV}     & 11.40    / 6.62                                 & 3.15 / 3.46                         & 1.28 / 2.22                         \\
    CurvilinearGANSpace (ours)                   & \hphantom{0}\underline{0.07} / \underline{0.35} & \underline{0.05} / \underline{0.62} & \underline{0.08} / \underline{0.55} \\
    \bottomrule
  \end{tabular}\\
  A: ``age'', G: ``gender'', R: ``race'' B: ``bangs'', P: ``pitch'', Y: ``yaw''.
  \vspace*{-2mm}
\end{table}

\subsection{Experimental Results}

\paragraph{Commutativity of Editing}

% We found that CelebA-HQ attributes classifier did not identify meaningful indices corresponding to beard and smile attributes for all editing methods; we summarized the numerical evaluations of the remaining attributes, where we set the normalized change amount $\tilde t=0.1$.

Table~\ref{tab:commutativity-stylegan2} shows the commutativity errors for CelebA-HQ+StyleGAN2 with $\tilde t=0.1$.
Those of LinearGANSpace and CurvilinearGANSpace were always less than 0.7 \%; even though they were not exactly zero due to numerical and rounding errors, the errors were negligible.
The errors of WarpedGANSpace were between 1.2 \% and 11.4 \%.
Therefore, as expected, the image editing by WarpedGANSpace is non-commutative, whereas that by LinearGANSpace and CurvilinearGANSpace is commutative.

We edited image attributes sequentially so that the total amount of change is zero and summarized the results in Fig.~\ref{fig:result-commutive}.
The images generated after sequential editing by LinearGANSpace or CurvilinearGANSpace look identical to the originals, which indicates that their image editing is commutative.
When editing a human face by WarpedGANSpace, the face's yaw rotation and image brightness were not restored.
Also for an AnimeFaces image, the hair color was not restored.
These results indicate that image editing by WarpedGANSpace is non-commutative.

A closer look at each edit reveals that the editing of a human face by LinearGANSpace does not properly edit the smile attribute, and the edit of yaw rotation changes the hairstyle as well.
When WarpedGANSpace edits the pitch or yaw rotation of the human face, it changes hair color, skin color, and brightness as well.
For the AnimeFaces image, the editing by LinearGANSpace is of inferior quality.
WarpedGANSpace's edits of the hair color and hair length change the face (i.e., identity).
CurvilinearGANSpace's editing is of excellent quality without severe side effects.

These results indicate that CurvilinearGANSpace provides commutative editing and significantly improves the disentanglement between attributes.
The training framework used~\cite{voynov2020unsupervised} leads the editing methods to learn disentanglement between attributes by classifying indices.
WarpedGANSpace takes advantage of nonlinearity to allow better editing; however, there is no mechanism to further facilitate disentanglement.
CurvilinearGANSpace assumes that attribute vector fields are locally linearly independent, and hence, always assigns a different direction to each attribute, which facilitates the disentanglement.
We describe disentanglement in the following section.

\begin{table}[t]
  \centering\footnotesize
  \tabcolsep  = 1.3mm
  \caption{Side Effect Errors [\%] of StyleGAN2. }
  \label{tab:sideeffect-stylegan2}
  \begin{tabular}{lcrrrrrr}
    \toprule
                                                 &                                     & \multicolumn{6}{c}{\textbf{Side Effect Errors} $l$ [\%]}                                                                                                                                                                      \\
    \cmidrule(rl){3-8}
                                                 & \!\!\!\!\textbf{Target} $k$\!\!\!\! & \multicolumn{1}{c}{\textbf{A}}                           & \multicolumn{1}{c}{\textbf{G}} & \multicolumn{1}{c}{\textbf{R}} & \multicolumn{1}{c}{\textbf{B}} & \multicolumn{1}{c}{\textbf{P}} & \multicolumn{1}{c}{\textbf{Y}} \\
    \midrule
    LinearGANSpace~\cite{voynov2020unsupervised} & A                                   & \underline{100}                                          & 59                             & 37                             & 63                             & 41                             & 61                             \\
                                                 & G                                   & 28                                                       & \underline{100}                & 16                             & 78                             & 20                             & 17                             \\
                                                 & R                                   & 61                                                       & 52                             & \underline{100}                & 71                             & 24                             & 19                             \\
                                                 & B                                   & \textbf{\textcolor{red}{175}}                            & \textbf{\textcolor{red}{172}}  & 78                             & \underline{100}                & 70                             & 64                             \\
                                                 & P                                   & 71                                                       & \textbf{\textcolor{red}{90}}   & 43                             & 76                             & \underline{100}                & 57                             \\
                                                 & Y                                   & 58                                                       & 55                             & 43                             & \textbf{\textcolor{red}{94}}   & 36                             & \underline{100}                \\
    \midrule
    WarpedGANSpace~\cite{Tzelepis_2021_ICCV}     & A                                   & \underline{100}                                          & 51                             & 63                             & \textbf{\textcolor{red}{111}}  & 59                             & 23                             \\
                                                 & G                                   & 75                                                       & \underline{100}                & \textbf{\textcolor{red}{94}}   & \textbf{\textcolor{red}{124}}  & \textbf{\textcolor{red}{236}}  & 57                             \\
                                                 & R                                   & 63                                                       & 64                             & \underline{100}                & \textbf{\textcolor{red}{131}}  & 73                             & 25                             \\
                                                 & B                                   & 23                                                       & 27                             & 22                             & \underline{100}                & 15                             & 21                             \\
                                                 & P                                   & 41                                                       & 44                             & 30                             & 80                             & \underline{100}                & 41                             \\
                                                 & Y                                   & 30                                                       & 30                             & 22                             & \textbf{\textcolor{red}{97}}   & 23                             & \underline{100}                \\
    \midrule
    CurvilinearGANSpace (ours)\!\!\!\!           & A                                   & \underline{100}                                          & 80                             & 45                             & \textbf{\textcolor{red}{137}}  & 60                             & 37                             \\
                                                 & G                                   & 62                                                       & \underline{100}                & 50                             & 84                             & 61                             & 40                             \\
                                                 & R                                   & 65                                                       & 56                             & \underline{100}                & 60                             & 37                             & 23                             \\
                                                 & B                                   & 40                                                       & 38                             & 15                             & \underline{100}                & 14                             & 19                             \\
                                                 & P                                   & 60                                                       & 52                             & 36                             & 76                             & \underline{100}                & 44                             \\
                                                 & Y                                   & 41                                                       & 62                             & 21                             & 79                             & 21                             & \underline{100}                \\
    \bottomrule
  \end{tabular}\\
  A                                            : ``age'', G      : ``gender'', R                  : ``race'' B                     : ``bangs'', P                   : ``pitch'' Y                    : ``yaw''.\\
  Severe side effects (more than 90 \%) are highlighted in bold red.
  \vspace*{-2mm}
\end{table}

\begin{table}[t]
  \centering\footnotesize
  \tabcolsep=1.1mm
  \caption{Identity Errors [\%] of StyleGAN2.}
  \label{tab:identity-stylegan2}
  \begin{tabular}{lrrrrrrr}
    \toprule
                                                                       & \multicolumn{1}{c}{\textbf{A}} & \multicolumn{1}{c}{\textbf{G}} & \multicolumn{1}{c}{\textbf{R}} & \multicolumn{1}{c}{\textbf{B}} & \multicolumn{1}{c}{\textbf{P}} & \multicolumn{1}{c}{\textbf{Y}} & \multicolumn{1}{c}{\textbf{Avg.}} \\
    \midrule
    \scalebox{0.95}[1.0]{LinearGANSpace}~\cite{voynov2020unsupervised} & \underline{26.1}               & \textbf{5.5}                   & \textbf{19.1}                  & 47.4                           & 26.4                           & 24.7                           & 29.9                              \\
    \scalebox{0.95}[1.0]{WarpedGANSpace}~\cite{Tzelepis_2021_ICCV}     & 27.6                           & 56.2                           & 33.6                           & \underline{6.3}                & \textbf{14.6}                  & \textbf{8.4}                   & \underline{29.3}                  \\
    \scalebox{0.95}[1.0]{CurvilinearGANSpace (ours)} \!\!\!            & \textbf{21.1}                  & \underline{15.4}               & \underline{25.3}               & \textbf{6.0}                   & \underline{18.9}               & \underline{9.6}                & \textbf{19.2}                     \\
    \bottomrule
  \end{tabular}\\
  A: ``age'', G: ``gender'', R: ``race'' B: ``bangs'', P: ``pitch'', Y: ``yaw'',\\ Avg.: average.
  \vspace*{-2mm}
\end{table}

\begin{figure}[t]
  \tabcolsep=.5mm
  \centering
  \begin{tabular}{c}
    \hspace*{0mm}\includegraphics[width=6.6cm]{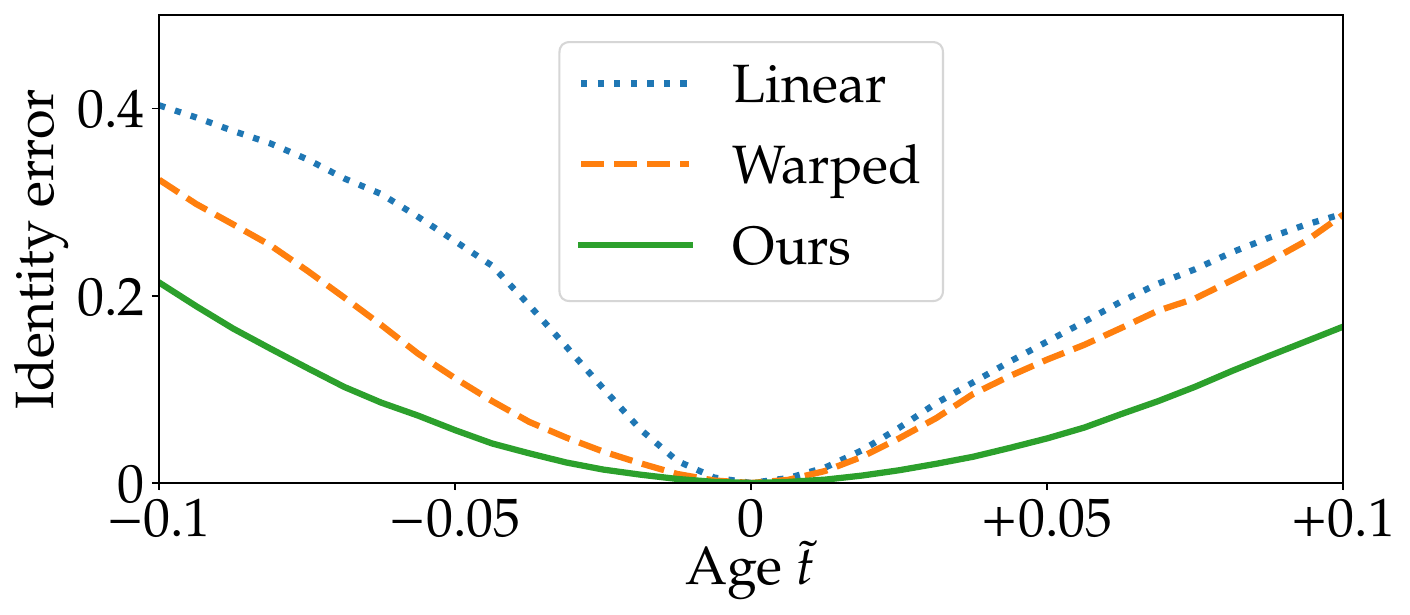} \\[-1mm]
    \begin{tabular}{cc}
      \rotatebox{90}{\footnotesize \ \ Linear} & \includegraphics[clip,width=6.3cm]{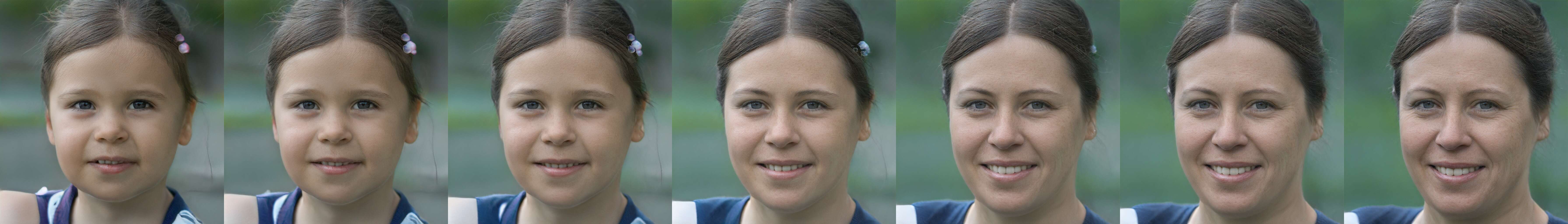}   \\[-.7mm]
      \rotatebox{90}{\footnotesize Warped}     & \includegraphics[clip,width=6.3cm]{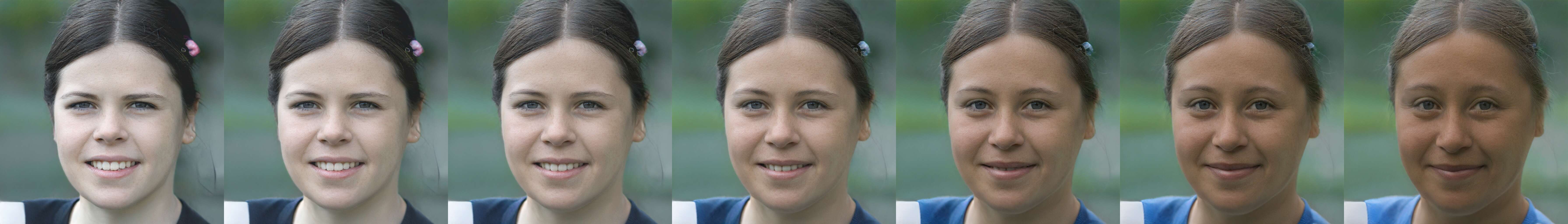}   \\[-.7mm]
      \rotatebox{90}{\footnotesize \ \ Ours}   & \includegraphics[clip,width=6.3cm]{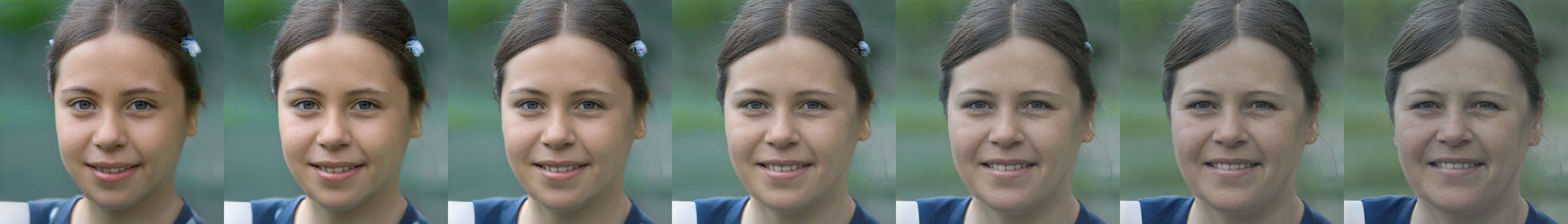} \\
    \end{tabular}
  \end{tabular}\\[2mm]
  \begin{tabular}{c}
    \hspace*{0mm}\includegraphics[width=6.6cm]{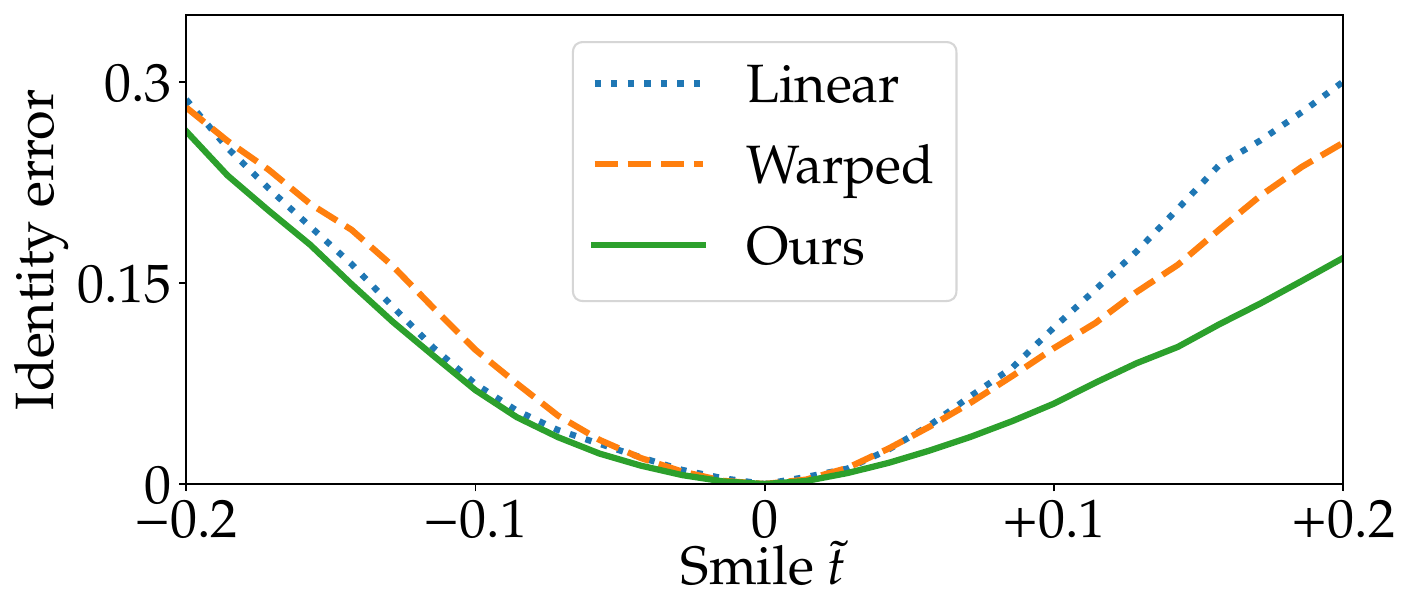} \\[-1mm]
    \begin{tabular}{cc}
      \rotatebox{90}{\footnotesize \ \ Linear} & \includegraphics[clip,width=6.3cm]{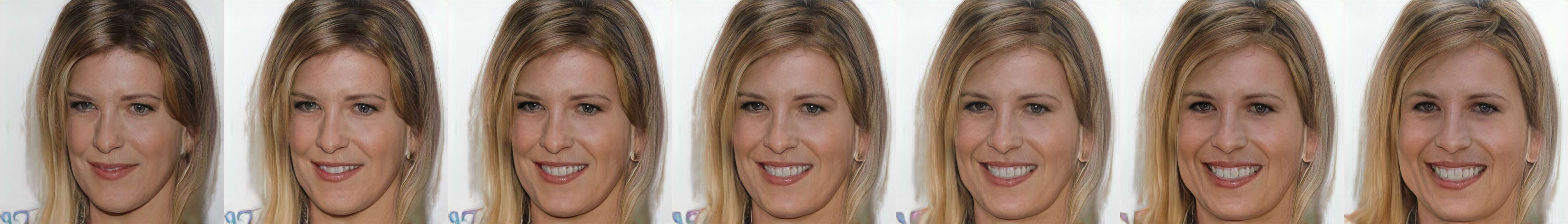}   \\[-.7mm]
      \rotatebox{90}{\footnotesize Warped}     & \includegraphics[clip,width=6.3cm]{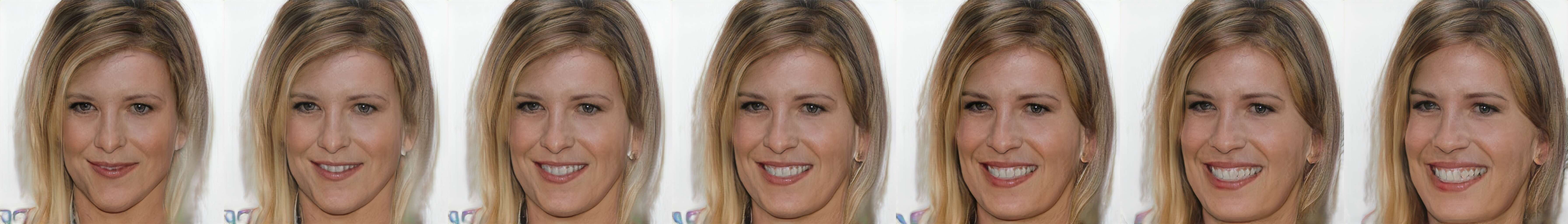}   \\[-.7mm]
      \rotatebox{90}{\footnotesize \ \ Ours}   & \includegraphics[clip,width=6.3cm]{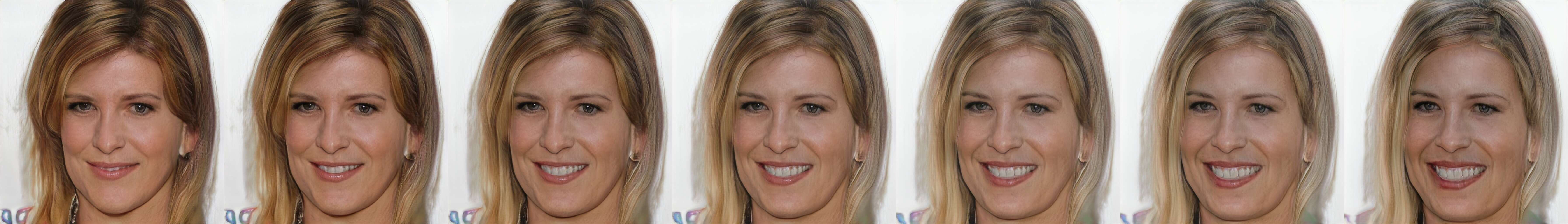} \\
    \end{tabular}
  \end{tabular}\\[-1mm]
  \caption{Identity errors when editing an attribute gradually.
    (top) Editing the age attribute of a StyleGAN2 image.
    (bottom) Editing the smile attribute of a ProgGAN image.
    Each row shows the results of LinearGANSpace, WarpedGANSpace, and CurvilinearGANSpace, from top to bottom.
  }
  \label{fig:identity}
  \vspace*{-2mm}
\end{figure}
\begin{figure*}[t]
  \centering\footnotesize
  \tabcolsep=0.3mm
  \begin{tabular}{cc@{\hspace{1.5mm}}c@{\hspace{1.5mm}}c@{\hspace{1.5mm}}c}
    \rotatebox{90}{\footnotesize Linear}            & \includegraphics[clip,width=5.5cm]{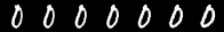}   & \includegraphics[clip,width=5.5cm]{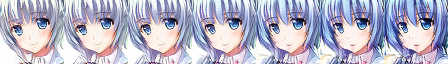}     & \includegraphics[clip,width=5.5cm]{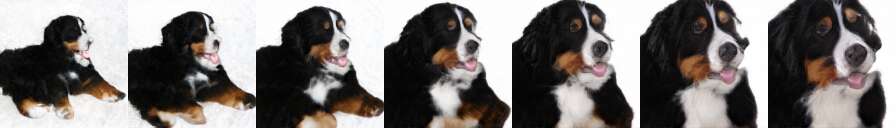}    \\[-0.7mm]
    \rotatebox{90}{\footnotesize \!\!\!\! Warped\!} & \includegraphics[clip,width=5.5cm]{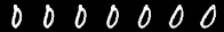}   & \includegraphics[clip,width=5.5cm]{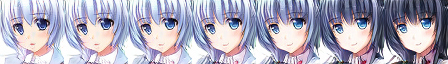}     & \includegraphics[clip,width=5.5cm]{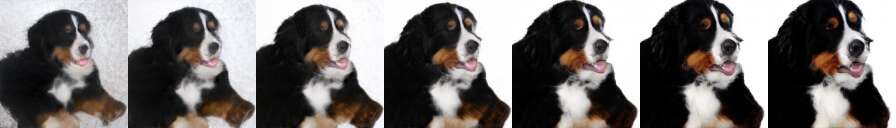}    \\[-0.7mm]
    \rotatebox{90}{\footnotesize \ Ours}            & \includegraphics[clip,width=5.5cm]{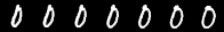} & \includegraphics[clip,width=5.5cm]{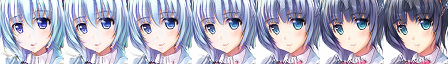}   & \includegraphics[clip,width=5.5cm]{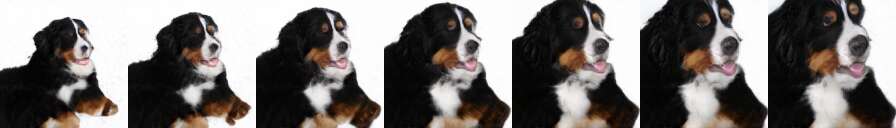}  \\[-0.7mm]
                                                    & (a) MNIST, ``width''.                                              & (c) AnimeFaces, ``hair color''.                                     & (e) ILSVRC, ``object size''.                                           \\[0.5mm]
    \rotatebox{90}{\footnotesize \ Linear}          & \includegraphics[clip,width=5.5cm]{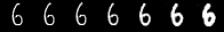}     & \includegraphics[clip,width=5.5cm]{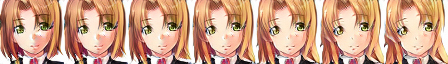}   & \includegraphics[clip,width=5.5cm]{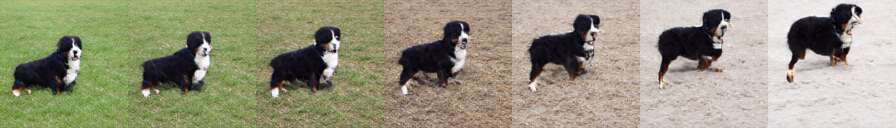}   \\[-0.7mm]
    \rotatebox{90}{\footnotesize \!\!\!\! Warped\!} & \includegraphics[clip,width=5.5cm]{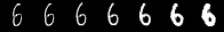}     & \includegraphics[clip,width=5.5cm]{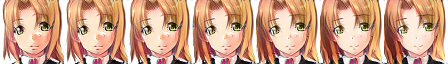}   & \includegraphics[clip,width=5.5cm]{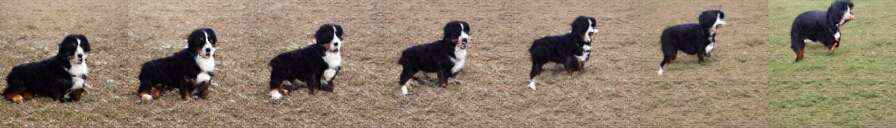}   \\[-0.7mm]
    \rotatebox{90}{\footnotesize \ Ours}            & \includegraphics[clip,width=5.5cm]{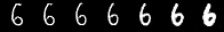}   & \includegraphics[clip,width=5.5cm]{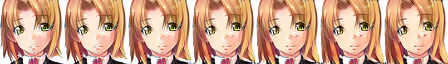} & \includegraphics[clip,width=5.5cm]{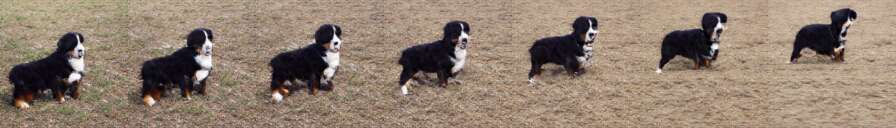} \\[-0.7mm]
                                                    & (b) MNIST, ``thickness''.                                          & (d) AnimeFaces, ``hair length''.                                    & (f) ILSVRC, ``vertical position''.                                     \\[0.5mm]
  \end{tabular}
  \begin{tabular}{cc@{\hspace{4.5mm}}c}
    \rotatebox{90}{\footnotesize \ Linear}          & \includegraphics[width=5.5cm]{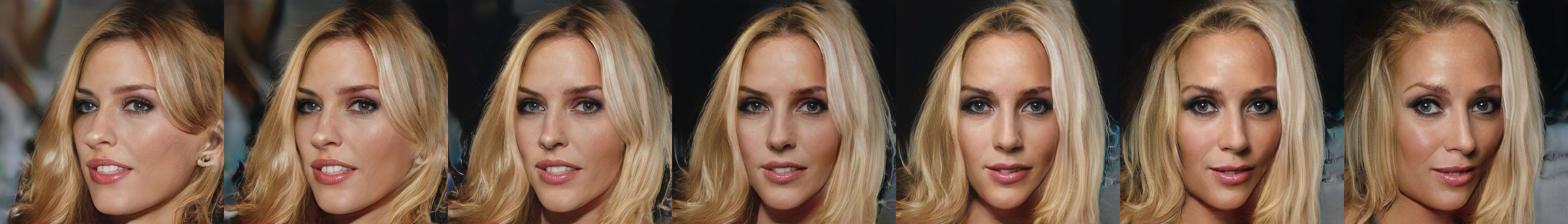}                           & \includegraphics[clip,width=7.2cm]{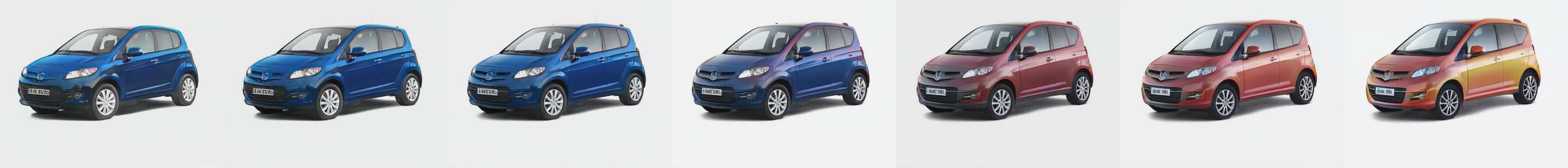}      \\[-.7mm]
    \rotatebox{90}{\footnotesize \!\!\!\! Warped\!} & \includegraphics[width=5.5cm]{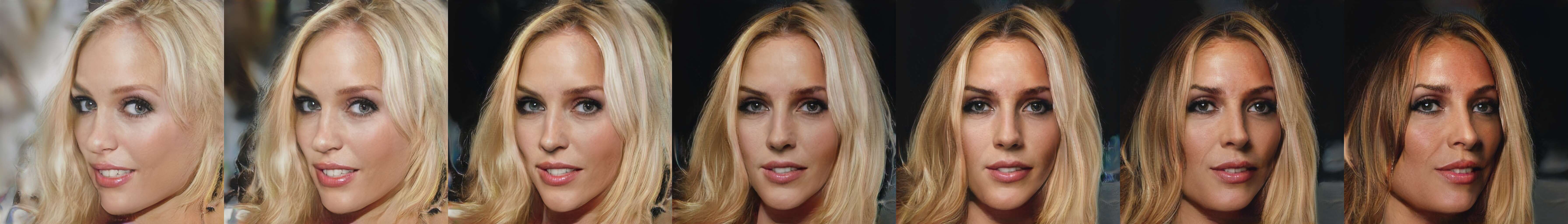}                           & \includegraphics[clip,width=7.2cm]{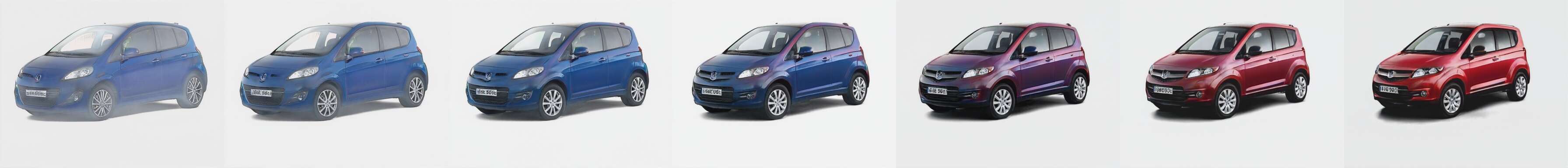}      \\[-.7mm]
    \rotatebox{90}{\footnotesize \ Ours}            & \includegraphics[width=5.5cm]{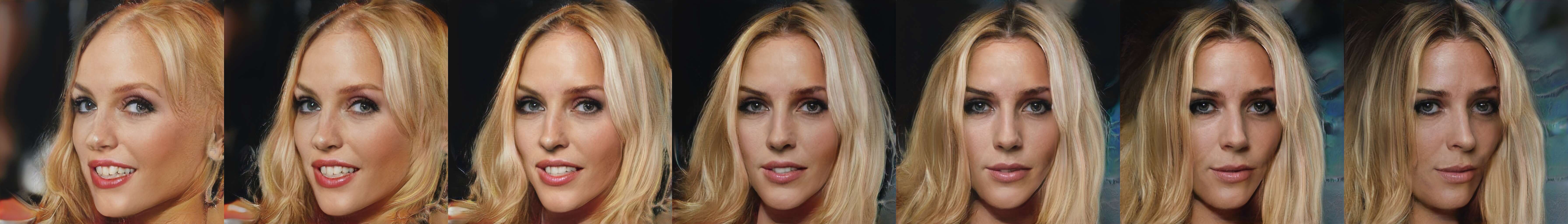}                         & \includegraphics[clip,width=7.2cm]{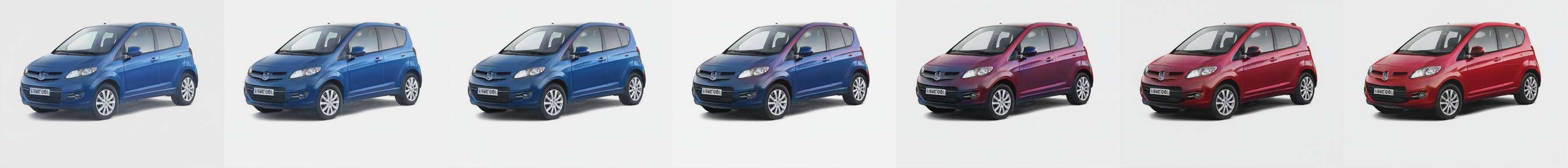}    \\[-1mm]
                                                    & {\tiny$-30\tcdegree$}\hspace{8.5mm} (g) ProgGAN, ``yaw''.\hspace{8.5mm}{\tiny$+30\tcdegree$} & (i) LSUN Car, ``color''.                                          \\[0.5mm]
    \rotatebox{90}{\footnotesize \ Linear}          & \includegraphics[width=5.5cm]{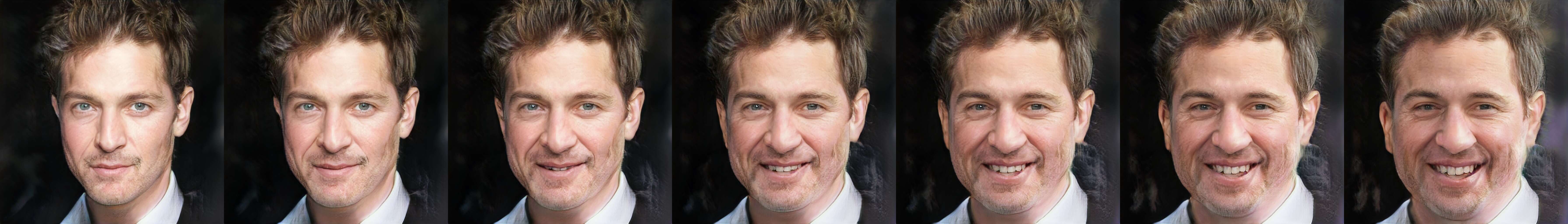}                                   & \includegraphics[clip,width=7.2cm]{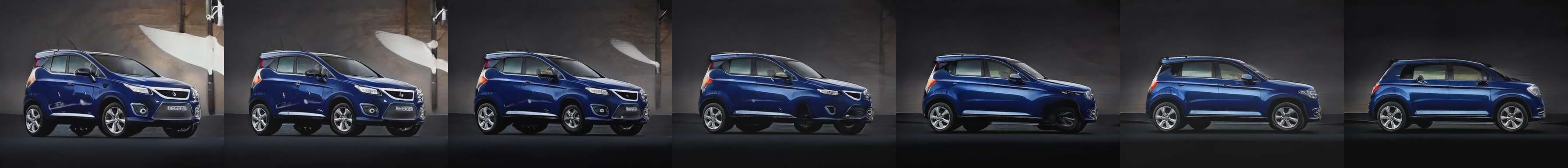}   \\[-0.7mm]
    \rotatebox{90}{\footnotesize \!\!\!\! Warped\!} & \includegraphics[width=5.5cm]{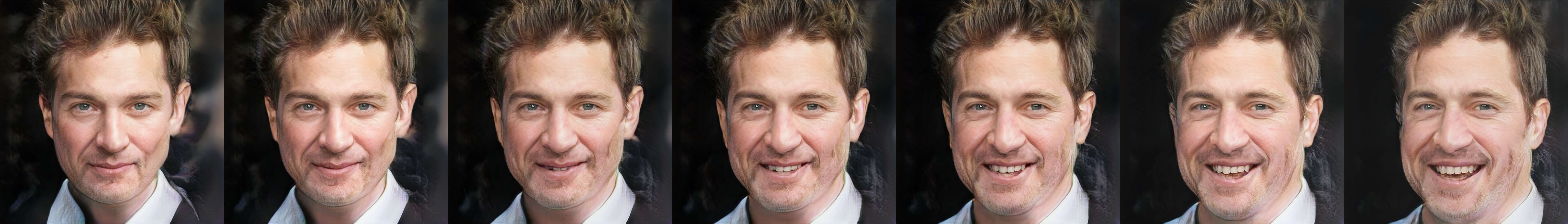}                                   & \includegraphics[clip,width=7.2cm]{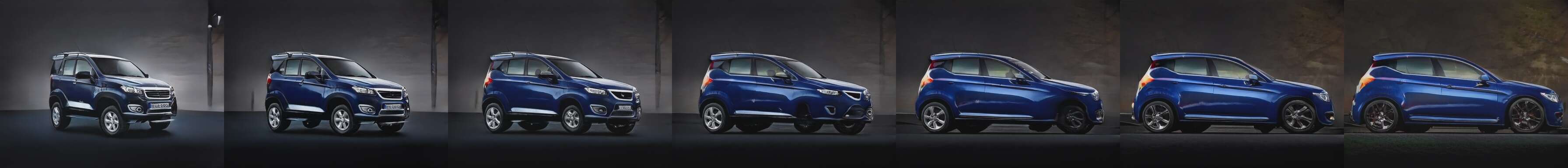}   \\[-0.7mm]
    \rotatebox{90}{\footnotesize \ Ours}            & \includegraphics[width=5.5cm]{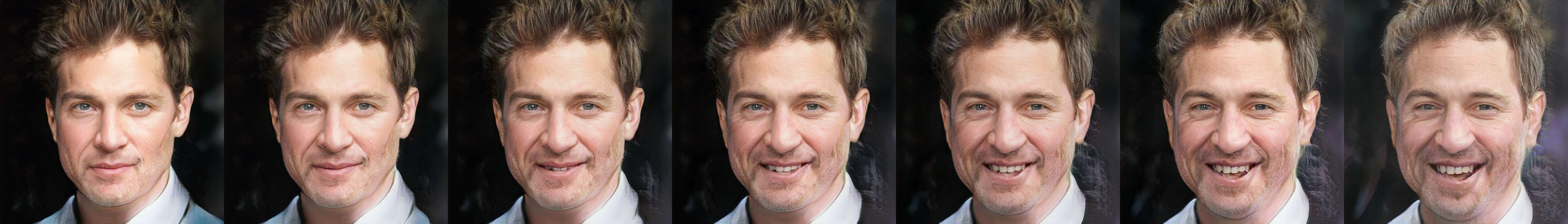}                              & \includegraphics[clip,width=7.2cm]{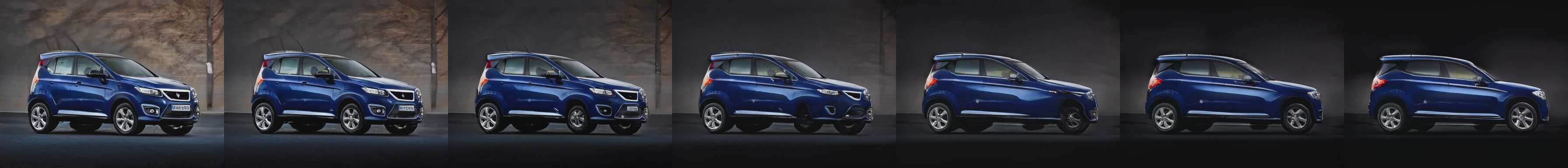} \\[-0.7mm]
                                                    & {\tiny$-0.2$}\hspace{8mm} (h) ProgGAN, ``smile''. \hspace{8mm}{\tiny$+0.2$}                  & (j) LSUN Car, ``rotation''.                                       \\[0.5mm]
  \end{tabular}\\
  \vspace*{-2mm}
  \caption{Visualization results.
    The models and edited attributes $k$ are shown below the panels.
    The image in the center is the original, the images on the right have attributes added, and the images on the left have attributes subtracted, as is the case with Fig.~\ref{fig:identity}.
  }\label{fig:visual-results}
  \vspace*{-2mm}
\end{figure*}

\paragraph{Disentanglement of Attributes}
Table~\ref{tab:sideeffect-stylegan2} shows the side effect errors.
We defined a ``severe side effect'' as a change in another attribute by 0.09 or more when editing a target attribute by $\tilde t=0.1$, and highlighted it in bold red.
All three editing methods confounded the age and bangs attributes (i.e., editing one impacted the other) owing to their high correlation, caused by unsupervised learning.
CurvilinearGANSpace has no other severe side effects, whereas LinearGANSpace and WarpedGANSpace have many severe side effects; for example, WarpedGANSpace's edit of the race attribute rather changes the bangs attribute.

Table~\ref{tab:identity-stylegan2} shows the identity errors.
CurvilinearGANSpace produced the lowest errors for two of the six attributes, the second lowest errors for the remaining, and the lowest average error.
LinearGANSpace and WarpedGANSpace produced severe identity errors in some cases.

These results indicate that only CurvilinearGANSpace selectively edits target attributes and preserves as much other information as possible, that is, it is of excellent quality for the disentanglement of attributes.

\paragraph{Visualization of Disentanglement}
We summarized the identity errors and generated images when editing attributes in Fig.~\ref{fig:identity}.
The larger the attribute editing, the greater the increase in identity error.
However, CurvilinearGANSpace has the lowest identity errors.
When WarpedGANSpace edited the age attribute of a StyleGAN2 image, it also altered the skin color, hair color, and facial expression.
LinearGANSpace also lost the identity.
WarpedGANSpace's edit of the smile attribute of a ProgGAN image altered the pitch and yaw rotations.
CurvilinearGANSpace can edit only specific attributes with a smaller loss of identity.

We summarized the results of other models when editing specific attributes in Fig.~\ref{fig:visual-results}.
The image editing by CurvilinearGANSpace was as intended with the least side effects; however, the image editing by LinearGANSpace or WarpedGANSpace exhibited severe side effects.
Panels (a) and (b) show the edits of the width of digit 0 and the thickness of digit 6 in the MNIST dataset, respectively.
LinearGANSpace and WarpedGANSpace additionally rotated the digits, while CurvilinearGANSpace maintained the original direction.
Panels (c) and (d) show the results of the AnimeFaces dataset.
The edit of the hair color by LinearGANSpace or WarpedGANSpace altered the face (i.e., loses the identity).
When LinearGANSpace and WarpedGANSpace lengthened the hair, they paradoxically reduced the face's shading.
Panels (e) and (f) demonstrate that, when LinearGANSpace and WarpedGANSpace enlarged or vertically moved dogs in photos, they changed the orientations and backgrounds.
Panel (g) shows that, when editing the yaw attribute, LinearGANSpace also edits the hairstyle, and WarpedGANSpace edits the skin color.
Panel (h) shows that WarpedGANSpace's edit of the smile attribute altered the pitch and yaw rotations, as in Fig.~\ref{fig:identity}.
Panels (i) and (j) show similar tendencies for LSUN Car.

Therefore, we conclude that the nonlinear and commutative nature of DeCurvEd contributes to the disentanglement between attributes and high-quality editing.
We also provide additional results in Appendix~\ref{appendix:more_results}.

\section{Conclusion}
This study proposed deep curvilinear editing (DeCurvEd), which defines a curvilinear coordinate on the latent space of generative models and edits images along axes of the coordinate.
DeCurvEd's edits of semantic attributes are theoretically nonlinear and commutative.
Combined with pre-trained GANs, we proposed CurvilinearGANSpace and experimentally demonstrated that it is superior to previous methods whose edits are linear or non-commutative in terms of the disentanglement between attributes and the preservation of identity.
Future work will focus on a combination of DeCurvEd with other deep generative models in supervised and unsupervised learning, such as Khrulkov et al.~\cite{Khrulkov2021LatentTV}.

\noindent\textbf{Limitations:}
Because DeCurvEd assumes a continuous change in attribute, it is unavailable for a discrete attribute, such as ``wearing sunglasses.''
A combination with discrete attributes remains a topic for future research.

\noindent\textbf{Acknowledgements:}
This work was supported by JST PRESTO (JPMJPR21C7), CREST (JPMJCR1914), and JSPS KAKENHI (19K20344, 19H04172), Japan.

  %%%%%%%%% REFERENCES
  {\small
    % \bibliographystyle{ieee_fullname}
    % \bibliography{egbib}

  }

\clearpage
\newpage
\clearpage
\newpage
\appendix

\renewcommand\thetable{A\arabic{table}}
\setcounter{table}{0}
\renewcommand\thefigure{A\arabic{figure}}
\setcounter{figure}{0}
\renewcommand\theequation{A\arabic{equation}}
\setcounter{equation}{0}

\section{Proofs of Remarks}\label{appendix:proof_of_remarks}

We took Theorems~\ref{theorem:commuting_vector_fields} and~\ref{theorem:commutativeVF} from Theorem 9.44 and 9.46 in the reference~\cite{Lee2012}.
Their proofs require many definitions and propositions that have not been directly used in this manuscript.
Hence, interested readers are referred to the reference~\cite{Lee2012}.
We provide the proofs of remarks as follows:

\begin{proof}[Proof of Remark \ref{remark:linear_is_commutative}]
  A method that assumes linear attribute arithmetic edits an attribute $k$ by adding an attribute vector $a_k$, which is independent of the position $z$, scaled by a change amount $t$, that is, $z+t\,a_k$.
  We can define a vector field $Z_k(z)\equiv a_k$ and the flow $\phi_k^t(z)=z+\int_0^t Z_k\mathrm{d}\tau=z+t\,a_k$.
  Therefore, it is a special case of a method that assumes attribute vector fields.
  Moreover, because it holds that $(\phi_l^s\circ\phi_k^t)(z)=(z+t\,a_k)+s\,a_l=(z+s\,a_l)+t\,a_k=(\phi_k^t\circ\phi_l^s)(z)$, its edits are commutative.
\end{proof}

\begin{proof}[Proof of Remark \ref{remark:vectorfields_are_not_commuting}]
  According to Theorem~\ref{theorem:commuting_vector_fields}, the flows of two vector fields do not commute in general.
  Edits by a method that assumes attribute vector fields follow their flows, which are not commuting in general.
\end{proof}

\begin{proof}[Proof of Remark \ref{remark:nonlinear_commutative}]
  While the flow $\psi_k$ on the Cartesianized latent space $\mathcal V$ is linear, the mapping $f$ can be nonlinear.
  As a result, the flow $\phi_k$ on the latent space $\mathcal Z$ can be nonlinear.
  For commutativity,
  \begin{equation}
    \begin{aligned}
      \phi_l^s\circ\phi_k^t
       & =f^{-1}\circ\psi_l^s\circ f\circ f^{-1}\circ\psi_k^t\circ f \\
       & =f^{-1}\circ\psi_l^s\circ\psi_k^t\circ f                    \\
       & =f^{-1}\circ\psi_k^t\circ\psi_l^s\circ f                    \\
       & =f^{-1}\circ\psi_k^t\circ f\circ f^{-1}\circ\psi_l^s\circ f \\
       & =\phi_k^t\circ\phi_l^s.
    \end{aligned}
  \end{equation}
\end{proof}
\begin{proof}[Proof of Remark \ref{remark:proposed_is_VF}]
  Given DeCurvEd, we can always define an attribute vector filed $Z_k$ on the latent space $\mathcal Z$ by pushforwarding the coordinate vector field $\tilde Z_k$ on the Cartesianized latent space $\mathcal V$; in particular,
  \begin{equation}
    \textstyle Z_k(z)=(f^{-1})_*(\tilde Z_k)= \frac{\partial f^{-1}(v)}{\partial v} e_k
  \end{equation}
  at point $z$ for $v=f(z)$.
  Hence, DeCurvEd always assumes a set of $N$ vector fields.
\end{proof}
\begin{proof}[Proof of Remark \ref{remark:linear_is_proposed}]
  Suppose the mapping $f$ of DeCurvEd is linear and non-degenerate (i.e., $f(z)= M z$ for a non-degenerate matrix $M$) and that the attribute vector $a_k$ on the latent space $\mathcal Z$ is defined as $a_k=M^{-1}e_k$.
  Then, it holds that
  \begin{equation}
    \begin{aligned}
      \phi_k^t(z)
       & =(f^{-1}\circ\psi_k^t\circ f)(x) \\
       & =M^{-1}\psi_k^t(Mz)              \\
       & =M^{-1}(t\,e_k+M z)              \\
       & =z+t\,a_k,                       \\
    \end{aligned}
  \end{equation}
  implying that an edit by a method that assumes linear attribute arithmetic is a special case of an edit by DeCurvEd.
\end{proof}

\section{Algorithms}\label{appendix:algorithm}
We summarize the edit by DeCurvEd in Algorithm \ref{alg:edit}.
We adopted the unsupervised training framework for GANs proposed by Voynov and Babenko~\cite{voynov2020unsupervised}; we summarize the framework in Algorithm~\ref{alg:training}.
The only difference from the original implementation is the latent variable manipulation and loss function at lines \ref{alg:training:edit} and \ref{alg:training:loss}, respectively.

For the change amount distribution $\mathcal P_\epsilon$, we first sampled the change amount $\epsilon'$ from a continuous uniform distribution $\mathcal U[-6,6]$.
Because the regression of very small changes does not contribute to proper learning, we rounded up small change amounts $\epsilon$; in particular, we considered $\epsilon=\mathrm{sign}(\epsilon')\cdot\max(|\epsilon'|,0.1)$.

\begin{algorithm}[h]
  \caption{Edit Attribute $k$}\label{alg:edit}
  \begin{algorithmic}[1]
    \renewcommand{\algorithmicrequire}{\textbf{Input:}}
    \renewcommand{\algorithmicensure}{\textbf{Output:}}
    \REQUIRE latent variable $z$,\! attribute index $k$,\! change amount $\epsilon$
    \ENSURE edited latent variable $z'$
    \STATE Obtain a mapped latent variable $v=f(z)$.
    \STATE Obtain the edited mapped latent variable $v'=v+\epsilon e_k$.
    %  by manipulating the mapped latent variable $v$.
    \STATE Obtain the edited latent variable $z'=f^{-1}(v')$.
  \end{algorithmic}
\end{algorithm}

\begin{algorithm}[h]
  \caption{Training CurvilinearGANSpace}\label{alg:training}
  \begin{algorithmic}[1]
    \renewcommand{\algorithmicrequire}{\textbf{Input:}}
    \renewcommand{\algorithmicensure}{\textbf{Output:}}
    \STATE Sample a latent variable $z$ from its prior $p(z)$.
    \STATE Sample an attribute index $k$ to be changed from the discrete uniform distribution $\mathcal U\{1,N'\}$.
    \STATE Sample a change amount $\epsilon$ from a continuous probability distribution $\mathcal P_\epsilon$.
    \STATE Edit latent variable $z$ using Algorithm~\ref{alg:edit}.\label{alg:training:edit}
    \STATE Generate a pair of images $x=G(z)$ and $x'=G(z')$ using the generator $G$.
    \STATE Feed the pair $(x,x')$ to the reconstructor $R$, and get two outputs $(\hat k,\hat\epsilon)$.
    \STATE Obtain the loss function that compares the outputs $(\hat k,\hat\epsilon)$ and the actual edit $(k,\epsilon)$.
    \STATE Train the mapping $f$ and the reconstructor $R$ jointly to minimize the loss function in Eq.~\eqref{eq:loss}.\label{alg:training:loss}
  \end{algorithmic}
\end{algorithm}

\section{Details of Experiments}
\subsection{Datasets and Backbones}\label{appendix:datasets}
In the experiments, we used the same combinations of the datasets, GANs, and reconstructors used in previous studies~\cite{voynov2020unsupervised,Tzelepis_2021_ICCV}.
GANs were pre-trained before being combined with the proposed method.
SNGANs were trained by us, and the other GANs were obtained from external repositories.
Reconstructors were trained jointly with the mapping $f$ from scratch.
We summarize them below.
\begin{enumerate}[itemsep=0ex]
  \item MNIST~\cite{LeCun2005TheMD} + Spectral Norm GAN (SNGAN)~\cite{Miyato2018} + LeNet~\cite{726791}.
        MNIST is a dataset of $32 \times 32$ monochrome images of hand-written digits.
        SNGAN had ResNet-like architecture composed of three residual blocks.
        The dimension number $N$ of the latent space $\mathcal Z$ is $N=128$.
  \item AnimeFaces dataset~\cite{Jin2017TowardsTH} + SNGAN + LeNet.
        AnimeFaces dataset contains $64 \times 64$ RGB images of cartoon characters' faces.
        SNGAN had ResNet-like architecture composed of four residual blocks with $N=128$.
  \item ILSVRC dataset~\cite{Deng2009ImageNetAL} + BigGAN~\cite{Brock2019LargeSG} + ResNet-18~\cite{He2016DeepRL}.
        ILSVRC dataset contains $128 \times 128$ RGB natural images.
        We obtained a pre-trained BigGAN with $N=120$.
  \item CelebA-HQ dataset~\cite{Liu2015DeepLF} + ProgGAN~\cite{Karras2018} + ResNet-18.
        CelebA-HQ dataset contains $1024 \times 1024$ RGB images of celebrities' faces.
        We obtained a pre-trained ProgGAN with $N=512$.
  \item CelebA-HQ dataset + StyleGAN2~\cite{Karras2019stylegan2} + ResNet-18.
        We obtained a pre-trained StyleGAN2 with $N=512$.
  \item LSUN Car dataset~\cite{KrambergerPotocnik2020} + StyleGAN2~\cite{Karras2019stylegan2} + ResNet-18.
        We obtained a pre-trained StyleGAN2 with $N=512$.
\end{enumerate}

\subsection{Normalizing Flow}\label{appendix:implementation}
For a smooth bijective mapping $f$, we employ a flow-based model~\cite{Kingma2018GlowGF}, namely a continuous normalizing flow (CNF)~\cite{chen2018neuralode,grathwohl2019ffjord}.
The CNF assumes an ordinary differential equation (ODE)
% \begin{equation}
$
  \textstyle \frac{\textrm{d}u}{\textrm{d}t} = g(u(t), t;\theta)
$
% \end{equation}
on the space equivalent to the latent space $\mathcal Z$, where $u$ denotes a state variable, $t$ denotes the time, and the function $g$ parameterized by $\theta$ maps the state $u$ to its time derivative.
Given an initial condition $u(0)=u_0$, the solution $u(t)$ is given by
% \begin{equation}
$
  \textstyle u(t)=u_0+\int_0^t g(u(\tau),\tau;\theta)\mathrm{d}\tau.
$
% \end{equation}
The function $g$ is modeled by a neural network.
We define mapping $f$ as the integration of the above ODE from $0$ to $T$, namely $f:z=u(0)\mapsto v=u(T)$.
One can regard the mapping $f$ to be parameterized by $\theta$.
Additionally, its inverse mapping $f^{-1}$ is defined by the integration from $T$ to $0$.
Owing to the characteristics of ODE, the mapping $f$ is differentiable and bijective.
In practice, a numerical integration (such as a Runge-Kutta method) is required to solve the above ODE; numerical errors are introduced in the mapping $f$, but they are negligible.
We used the Dormand-Prince method to integrate the ODE for $T=0.1$.
For CurvilinearGANSpace, the log-determinant $\log\det\frac{\partial f}{\partial z}$ of the Jacobian of the mapping $f$ is stochastically obtainable using Hutchinson's estimator~\cite{chen2018neuralode,grathwohl2019ffjord}.

The CNF is guaranteed to be bijective, and serves as a universal approximator for smooth bijections, as proven in~\cite{Teshima2020a}.
Hence, DeCurvEd's editing is guaranteed to be commutative at the design stage, not trained to be commutative.
In practice, numerical errors during numerical integration cause a slight increase in the commutative error, but it remains negligible, as shown in Tables \ref{tab:commutativity-stylegan2}, \ref{tab:commutativity-appendix-proggan}, and \ref{tab:commutativity-appendix-stylegan2}.
Note that other normalizing flows are available~\cite{Kingma2018GlowGF}.

\begin{table*}[t]
  \centering\footnotesize
  \tabcolsep=1.5mm
  \caption{Commutativity Errors [\%] of ProgGAN.}
  \label{tab:commutativity-appendix-proggan}
  \begin{tabular}{lccc}
    \toprule
                                                 & \multicolumn{3}{c}{\textbf{ProgGAN}}                                                                                                                   \\
    \cmidrule(rl){2-4}
                                                 & \multicolumn{1}{c}{\textbf{S+Y}}     & \multicolumn{1}{c}{\textbf{B+P}}    & \multicolumn{1}{c}{\textbf{S+B+Y+P}}                                      \\
    \midrule
    LinearGANSpace~\cite{voynov2020unsupervised} & \textbf{0.09} / \textbf{0.12}        & \textbf{0.09}/ \textbf{0.13}        & \textbf{0.08} / \textbf{0.07} / \textbf{0.12} / \textbf{0.20}             \\
    WarpedGANSpace~\cite{Tzelepis_2021_ICCV}     & 5.86    / 1.97                       & 5.87 / 2.49                         & 1.51 / 7.80 / 3.00 / 2.08                                                 \\
    CurvilinearGANSpace (ours)                   & \underline{0.32} / \underline{0.44}  & \underline{0.24} / \underline{0.59} & \underline{0.22} / \underline{0.25} / \underline{0.64} / \underline{0.51} \\
    \bottomrule
  \end{tabular}\\
  S: ``Smile'', B: ``bangs'', P: ``pitch'', Y: ``yaw''.
  % \end{table*}
  \vspace*{5mm}
  % \begin{table*}[t]
  \centering\footnotesize
  \tabcolsep=1.5mm
  \caption{Commutativity Errors [\%] of StyleGAN2.}
  \label{tab:commutativity-appendix-stylegan2}
  \begin{tabular}{lcccccc}
    \toprule
                                                 & \multicolumn{3}{c}{\textbf{StyleGAN2}}                                                                                                                                                                                           \\
    \cmidrule(rl){2-4}
                                                 & \multicolumn{1}{c}{\textbf{G+B+Y}}                     & \multicolumn{1}{c}{\textbf{A+R+P}}                     & \multicolumn{1}{c}{\textbf{A+B+G+R+Y+P}}                                                                       \\
    \midrule
    LinearGANSpace~\cite{voynov2020unsupervised} & \textbf{0.04} / \textbf{0.02} / \textbf{0.21}          & \textbf{0.01} / \textbf{0.01} / \textbf{0.16}          & \textbf{0.02} / \textbf{0.02} / \textbf{0.06} / \textbf{0.02} / \textbf{0.12} / \textbf{0.45}                  \\
    WarpedGANSpace~\cite{Tzelepis_2021_ICCV}     & 3.58 / 1.05 / 8.54                                     & 3.77 / 3.28 / 3.33                                     & 9.48 / 1.71 / 7.43 / 1.19 / 6.90 / 6.52                                                                        \\
    CurvilinearGANSpace (ours)                   & \underline{0.23} / \underline{0.07} / \underline{0.51} & \underline{0.09} / \underline{0.07} / \underline{0.90} & \underline{0.06} / \underline{0.03} / \underline{0.27}/ \underline{0.10} / \underline{0.89} / \underline{0.60} \\
    \bottomrule
  \end{tabular}\\
  A: ``age'', G: ``gender'', R: ``race'', B: ``bangs'', P: ``pitch'', Y: ``yaw''.\\
  % \end{table*}
  \vspace*{5mm}
  % \begin{figure*}[t]
  \centering
  \tabcolsep=.5mm
  \begin{minipage}{81mm}
    \begin{tabular}{cl|c}
      \rotatebox{90}{\footnotesize \ \ Linear} & \includegraphics[clip, width=6.3cm]{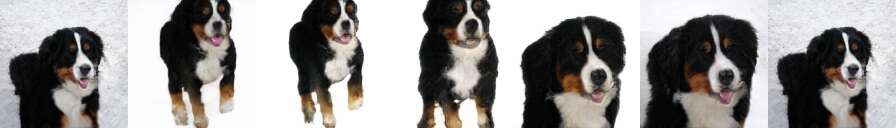}                                                               & \includegraphics[clip, width=0.9cm]{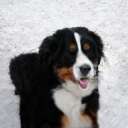} \\[-.8mm]
      \rotatebox{90}{\footnotesize Warped}     & \includegraphics[clip, width=6.3cm]{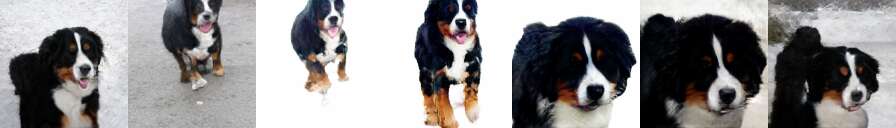}                                                               & \includegraphics[clip, width=0.9cm]{./fig/commutative_biggan1_original.jpg} \\[-.8mm]
      \rotatebox{90}{\footnotesize \ \ \ Ours} & \includegraphics[clip, width=6.3cm]{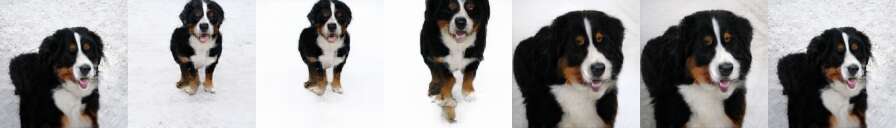}                                                             & \includegraphics[clip, width=0.9cm]{./fig/commutative_biggan1_original.jpg} \\[-1mm]
                                               & \footnotesize \hspace{3mm}O\hspace{5.5mm}$+$V\hspace{5.0mm}$-$B\hspace{5.0mm}$+$S\hspace{5.0mm}$-$V\hspace{5.0mm}$+$B\hspace{5.0mm}$-$S & \footnotesize O
    \end{tabular}
  \end{minipage}
  \begin{minipage}{81mm}
    \begin{tabular}{cl|c}
      \rotatebox{90}{\footnotesize \ \ Linear} & \includegraphics[clip, width=6.3cm]{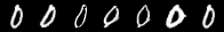}                                                                & \includegraphics[clip, width=0.9cm]{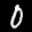} \\[-.8mm]
      \rotatebox{90}{\footnotesize Warped}     & \includegraphics[clip, width=6.3cm]{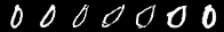}                                                                & \includegraphics[clip, width=0.9cm]{./fig/commutative_mnist0_original.jpg} \\[-.8mm]
      \rotatebox{90}{\footnotesize \ \ \ Ours} & \includegraphics[clip, width=6.3cm]{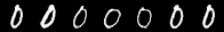}                                                              & \includegraphics[clip, width=0.9cm]{./fig/commutative_mnist0_original.jpg} \\[-1mm]
                                               & \footnotesize \hspace{3mm}O\hspace{5.5mm}$+$A\hspace{5.0mm}$-$T\hspace{5.0mm}$+$W\hspace{5.0mm}$-$A\hspace{5.0mm}$+$T\hspace{5.0mm}$-$W & \footnotesize O
    \end{tabular}
  \end{minipage}
  % \vspace*{-3mm}
  \captionof{figure}{Results of sequential editing of attributes from left to right.
    (left) ILSVRC+BigGAN.
    (right) MNIST+SNGAN.
    Each row shows the results of LinearGANSpace, WarpedGANSpace, and CurvilinearGANSpace, from top to bottom.
    The signs $+$ and $-$ denote the addition and the subtraction of the corresponding attributes, respectively.
    O: original, V: ``vertical position'', B: ``background'', S: ``object size''.
    A: ``angle'', T: ``thickness'', W: ``width''.
  }
  \label{fig:result-commutive-appendix}
  % \vspace*{-2mm}
  % \end{figure*}
\end{table*}

\subsection{Evaluation Metrics}\label{appendix:metrics}
\paragraph{Index Identification}
This process adopts the official implementation of WarpedGANSpace~\cite{Tzelepis_2021_ICCV}.
% We considered prepared latent variables $z$, edited the index $k$ from -3 to 3 in increments of 0.15, measured the attribute scores of generated images by the attribute predictors, and obtained the covariances between the change amount $t$ and the measured attribute scores.
We considered prepared latent variables $z$, edited the index $k$ by $t\in[-\tau,\tau]$ in increments of $\delta$, measured the attribute scores of generated images by the attribute predictors, and obtained the covariances between the change amount $t$ and the measured attribute scores.
$\tau$ and $\delta$ were set to $\tau=3$ and $\delta=0.15$ for StyleGAN2 and $\tau=4.5$ and $\delta=0.15$ for ProgGAN.
We selected index $k$ with the largest covariance as the one corresponding to that attribute.
Note that the original manuscript~\cite{Tzelepis_2021_ICCV} suggests using correlation; however, the implementation actually uses covariance.

\paragraph{Normalization}
We sampled 100 latent variables $z$, edited attribute $k$ by $t$, and obtained the edited latent variables $z'=\phi_k^t(z)$.
We generated the original $x=G(z)$ and edited $x'=G(z')$ images.
Using a separate attribute predictor $A_k$, we obtained the change in the attribute score in the image space $\mathcal X$, that is, $A_k(x')-A_k(x)$.
We obtained the average change $\mathbb E_z[A_k(G(\phi_k^t(z)))-A_k(G(z))]$ of the measured attribute score.
We identified the change amount $t$ in the latent space $\mathcal Z$ with which the average change was 5 degrees for the pitch and yaw attributes, and 0.1 for other attributes.
We normalized the change amount $t$ as $\tilde t=0.1$ for each attribute and method separately.

\paragraph{Commutativity Error}
Intuitively, \emph{commutativity error} is the error when edits of two attributes $k$ and $l$ are applied in reversed orders.
We defined it as follows:
Obtain a latent variable $z$, and edit attributes $k$ and $l$ by amounts $t$ and $s$ of latent variable $z$ in both orders; namely, obtain two latent variables $z_1=\phi_l^s(\phi_k^t(z))$ and $z_2=\phi_k^t(\phi_l^s(z))$.
Then, generate images $x_1=G(z_1)$ and $x_2=G(z_2)$, and evaluate the attributes scores of the generated images $x_1$ and $x_2$ by separate attribute predictors $A_k$ and $A_l$.
The commutativity error for attribute $k$ is the absolute difference $|A_k(x_1)-A_k(x_2)|$ in the attribute scores $A_k(x_1)$ and $A_k(x_2)$.
We obtained the errors for attributes $k$ and $l$; namely
\begin{equation}
  \begin{aligned}
     & |A_k(G(\phi_l^s(\phi_k^t(z))))-A_k(G(\phi_k^t(\phi_l^s(z))))|, \\
     & |A_l(G(\phi_l^s(\phi_k^t(z))))-A_l(G(\phi_k^t(\phi_l^s(z))))|.
  \end{aligned}
\end{equation}
We set the change amount to $\tilde t=\tilde s=0.1$ in the normalized scale.
This error vanishes if edits of attributes $k$ and $l$ are commutative.
For over two attributes, we obtained the difference in attribute score between edited results in the given order and in the reverse order.

\paragraph{Side Effect Error}
We defined the \emph{side effect error} as follows:
Obtain a latent variable $z$, and edit attribute $k$ by $t$, obtaining $z'=\phi_k^t(z)$.
Then, measure the difference $|A_l(x)-A_l(x')|$ in the score of other attribute $l$ between generated images $x=G(z)$ and $x'=G(z')$, and normalize it by that for the target attribute $k$; namely
\begin{equation}
  \frac{|A_l(G(z))-A_l(G(\phi_k^t(z)))|}{|A_k(G(z))-A_k(G(\phi_k^t(z)))|}
\end{equation}
We set the change amount to $\tilde t=0.1$ in the normalized scale.
This error vanishes if the edit of attribute $k$ has no side effect on attribute $l$.

\paragraph{Identity Error}
We defined the \emph{identity error} as follows:
Obtain a latent variable $z$, and edit attribute $k$ by $t$, obtaining $z'=\phi_k^t(z)$.
Then, evaluate the identity score $I(x,x')$ between the generated images $x=G(z)$ and $x'=G(z')$.
The identity error is defined as 1.0 minus the identity score; namely,
\begin{equation}
  1-I(G(z),G(\phi_k^t(z))).
\end{equation}
We set the change amount to $\tilde t=0.1$ in the normalized scale.
For more than two attributes, we also obtained 1.0 minus the identity score between the original and edited images.

\begin{figure*}[t]
  \centering\footnotesize
  \tabcolsep=0.3mm
  \begin{tabular}{c@{\hspace{1.5mm}}c@{\hspace{1.5mm}}c}
    \includegraphics[clip,width=5.5cm]{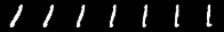} & \includegraphics[clip,width=5.5cm]{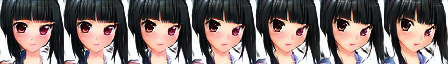} & \includegraphics[clip,width=5.5cm]{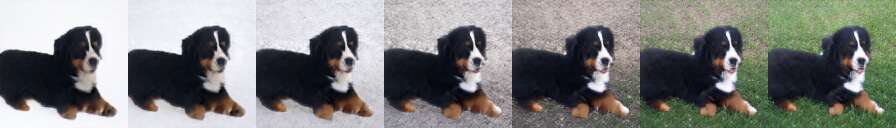} \\[-0.7mm]
    (a) MNIST, ``rotation''.                                              & (d) AnimeFaces, ``rotation''.                                         & (g) ILSVRC, ``background''.                                                 \\[1mm]
    \includegraphics[clip,width=5.5cm]{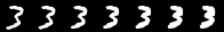}     & \includegraphics[clip,width=5.5cm]{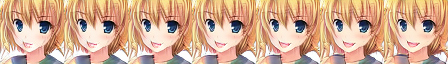}  & \includegraphics[clip,width=5.5cm]{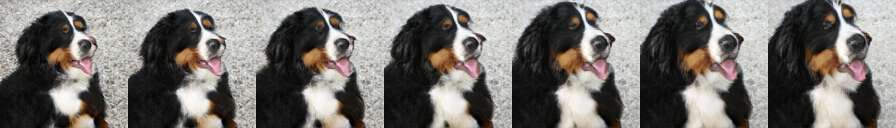}      \\[-0.7mm]
    (b) MNIST, ``thickness''.                                             & (e) AnimeFaces, ``hair length''.                                      & (h) ILSVRC, ``object size''.                                                \\[1mm]
    \includegraphics[clip,width=5.5cm]{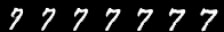}   & \includegraphics[clip,width=5.5cm]{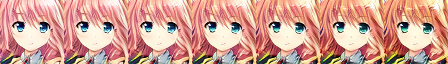}    & \includegraphics[clip,width=5.5cm]{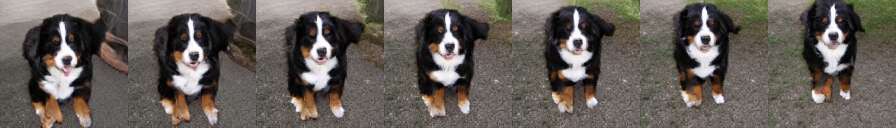}     \\[-0.7mm]
    (c) MNIST, ``width''.                                                 & (f) AnimeFaces, ``hair color''.                                       & (i) ILSVRC, ``vertical position''.                                          \\[1mm]
  \end{tabular}
  \begin{tabular}{c@{\hspace{3mm}}c}
    \includegraphics[clip,width=8cm]{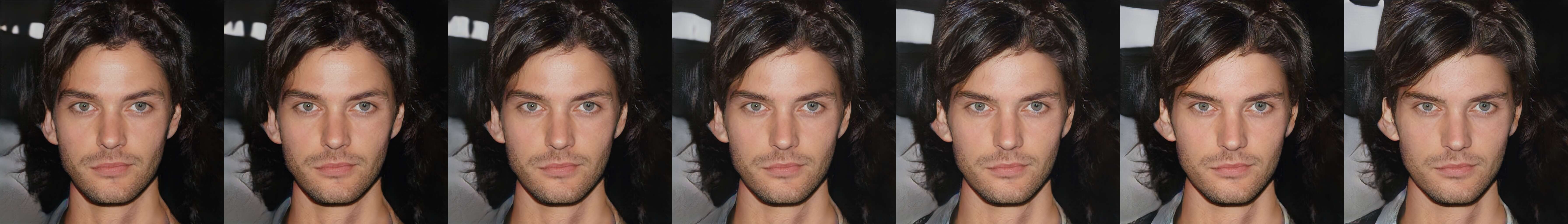}   & \includegraphics[clip,width=8cm]{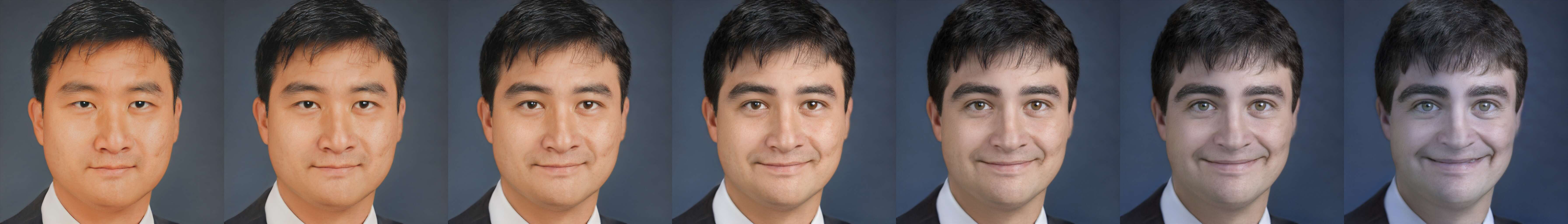}  \\[-1mm]
    $-0.1$\hspace{30.5mm}$0$\hspace{30.5mm}$+0.1$                             & $-0.1$\hspace{30.5mm}$0$\hspace{30.5mm}$+0.1$                             \\[-.5mm]
    (j) ProgGAN, ``bangs''.                                                   & (m) StyleGAN2, ``race''.                                                  \\[1mm]
    \includegraphics[clip,width=8cm]{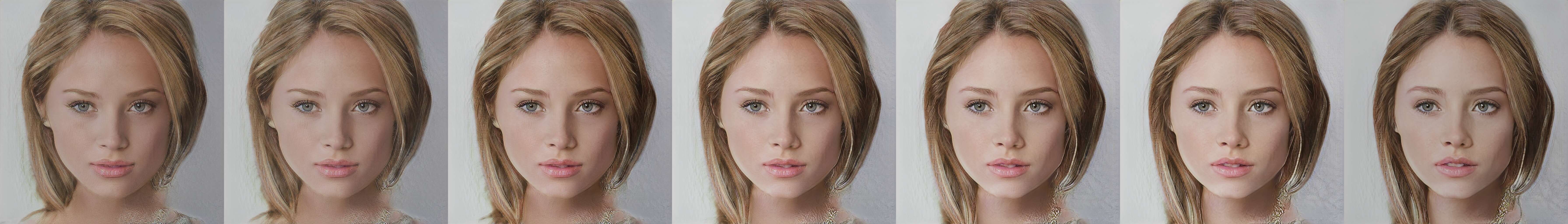} & \includegraphics[clip,width=8cm]{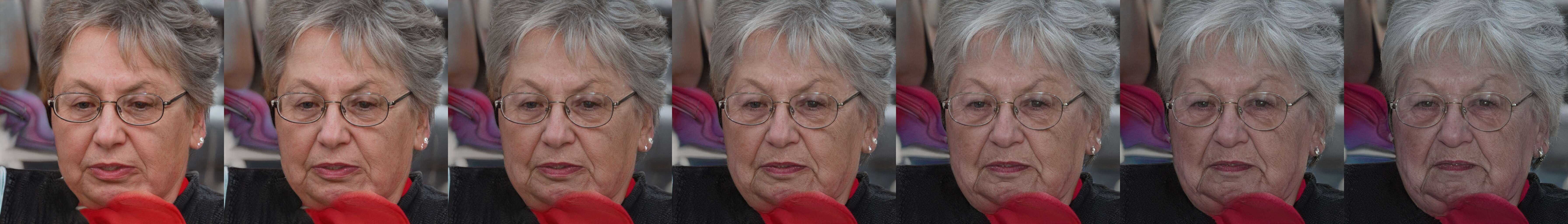} \\[-1mm]
    $-5\tcdegree$\hspace{30.5mm}$0$\hspace{30.5mm}$+5\tcdegree$               & $-0.1$\hspace{30.5mm}$0$\hspace{30.5mm}$+0.1$                             \\[-.5mm]
    (k) ProgGAN, ``pitch''.                                                   & (n) StyleGAN2, ``age''.                                                   \\[1mm]
    \includegraphics[clip,width=8cm]{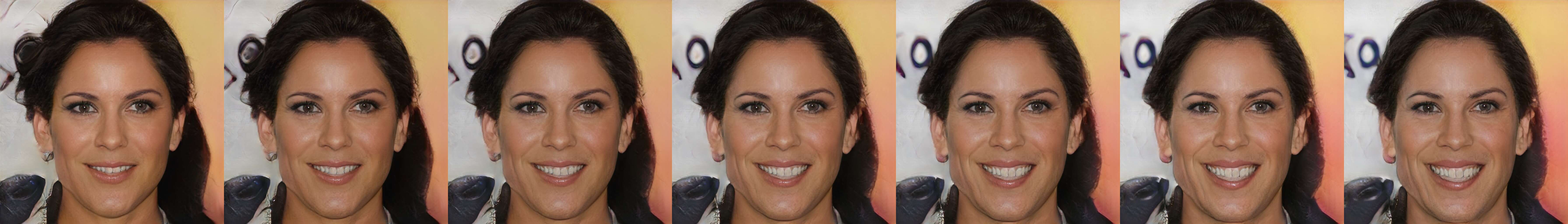} & \includegraphics[clip,width=8cm]{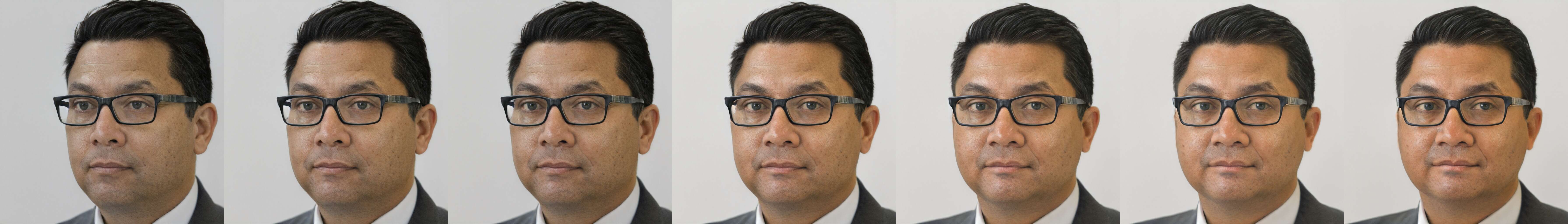}   \\[-1mm]
    $-0.1$\hspace{30.5mm}$0$\hspace{30.5mm}$+0.1$                             & $-5\tcdegree$\hspace{30.5mm}$0$\hspace{30.5mm}$+5\tcdegree$               \\[-.5mm]
    (l) ProgGAN, ``smile''.                                                   & (o) StyleGAN2, ``yaw''.
  \end{tabular}\\
  \vspace*{-2mm}
  \caption{Visualization results of CurvilinearGANSpace.
    The models and edited attributes $k$ are shown below the panels.
    The image in the center is the original, the images on the right have attributes added, and the images on the left have attributes subtracted, in the same way as Figs.~\ref{fig:identity} and \ref{fig:visual-results}.}
  \label{fig:proposed-visual-appendix}
\end{figure*}

\section{Additional Results}\label{appendix:more_results}
\subsection{Commutativity}
In this section, we provide additional results for demonstrating the commutativity of image editing methods.
In a way similar to Table~\ref{tab:commutativity-stylegan2}, Tables~\ref{tab:commutativity-appendix-proggan} and~\ref{tab:commutativity-appendix-stylegan2} show the commutativity errors.
For any combination of attributes, the errors of LinearGANSpace and CurvilinearGANSpace were always less than 0.9 \%, whereas those of WarpedGANSpace varied between 1.0 \% and 9.5 \%.

Following Fig.~\ref{fig:result-commutive}, we edited image attributes sequentially so that the total amount of change is zero and summarized the results in Fig.~\ref{fig:result-commutive-appendix}.
When using LinearGANSpace or CurvilinearGANSpace, the images returned to their original states.
WarpedGANSpace did not restore the original images; the position and background of the dog were not restored, and the digit was thickened.

These results also demonstrate that the image editing by LinearGANSpace and CurvilinearGANSpace is commutative and that by WarpedGANSpace is non-commutative.

\subsection{More Visualization}
We provide further visualization results of CurvilinearGANSpace in Fig.~\ref{fig:proposed-visual-appendix}, demonstrating that CurvilinearGANSpace identified and edited various attributes without severe side effects.

\end{document}